\PassOptionsToPackage{hidelinks}{hyperref}
\documentclass[preprint,authoryear,12pt]{elsarticle}

\usepackage{lmodern}
\usepackage[T1]{fontenc}
\usepackage{amsmath,amssymb,amsfonts}
\usepackage{booktabs}
\usepackage[section]{placeins}
\usepackage{enumitem}
\usepackage{microtype}
\usepackage{url}

\graphicspath{{figures/}}

\newcommand{\code}[1]{\texttt{#1}}
\makeatletter
\def\ps@pprintTitle{%
  \let\@oddhead\@empty
  \let\@evenhead\@empty
  \def\@oddfoot{\reset@font\hfil\thepage\hfil}%
  \let\@evenfoot\@oddfoot}
\makeatother

\begin{document}

\begin{frontmatter}

\title{Global tree forecasters collapse at the hierarchical aggregate:
a five-panel failure characterization}

\author[fn]{Md Rezwanul Islam\corref{cor1}}
\ead{rezwanul.islam@fieldnation.com}
\author[fn]{Wael Mohammed}
\ead{wael.mohammed@fieldnation.com}
\ead[url]{https://orcid.org/0009-0005-4211-8976}
\affiliation[fn]{organization={Field Nation LLC},
                 city={Minneapolis}, state={MN}, country={USA}}
\cortext[cor1]{Corresponding author. ORCID:
\texttt{https://orcid.org/0009-0002-4100-3796}.}

\begin{abstract}
Global forecasting models pool many series and learn one shared
function. Gradient-boosted trees are their most common form. We
measure a failure of this design that has not, to our knowledge, been
documented. Train a global tree on the individual series of a
hierarchy, then ask it for the hierarchical aggregate. The aggregate
sits far outside the model's training range, and the forecast
collapses. The model under-predicts the total by $30$--$50\times$ in
our production deployment, and by up to $496\times$ in a public M5
reconstruction. The mechanism is known: beyond its training range, a tree predicts
a constant. It surfaces at the aggregate because the total dwarfs every training
series. The cure is not new. Per-series scaling, the preprocessing
step that Montero-Manso and Hyndman (2021) recommend, prevents the
collapse. So do a weighted aggregate-level training row and seasonal
differencing. Our contribution is the characterization. The collapse
reproduces on five panels: a production business-to-business
marketplace, a synthetic hierarchy, M5, Australian Tourism, and a
public business-buyer panel. It holds on three tree libraries, is
invariant across training seeds, and is statistically significant.
Its onset is immediate and tracks a simple support bound: a scale
gap of only $1.15\times$ already costs a third of the total. No
standard configuration change prevents it: pooling every hierarchy
level into training fails at scale, and the one knob that fits
linear models in the leaves softens it without curing it. Rolling the forecasts
forward recursively separates the cures: the aggregate-row cure
re-collapses, per-series scaling degrades but stays low, and only
seasonal differencing keeps its one-step accuracy unchanged. We
close with a three-step procedure for diagnosing and
preventing the failure in deployed systems.
\end{abstract}


\end{frontmatter}

\section{Introduction}
\label{sec:intro}

Global forecasting models pool many series and learn one shared
function. They dominate applied work on large panels, and
gradient-boosted trees are the most common global model in practice
\citep{januschowski2020criteria,makridakis2022m5}. The published
evidence favors them. A simulation and benchmarking literature maps
when cross-learning helps
\citep{hewamalage2022global,bandara2020clustering,godahewa2021monash},
and \citet{montero2021globality} give theory for
why one global model can match or beat per-series local models.
Recent work reports a global LightGBM accurate across the levels of a
hierarchy \citep{zhao2024localglobal}. The same literature also
supplies the remedy this paper turns on. \citet{montero2021globality}
recommend, and apply, per-series scaling as a preprocessing step, so
that all series reach the model on a comparable scale. That cure is
established prior art. Nothing in this paper claims it.

What has not been documented is the failure the cure prevents. This
paper measures it.

\paragraph{The failure} Take a global tree with absolute lag
features, trained on the individual series of a hierarchy. Ask it to
forecast the hierarchical aggregate. It collapses. Our production
panel is monthly gross transaction value (GTV) from a
business-to-business service marketplace. There, the aggregate lag-1
input is about $1{,}088\times$ the median per-buyer lag-1 input. That
is three orders of magnitude above the median training input, and
altogether outside the model's training support---the range of
values seen in training.
Trees extrapolate as constants beyond their observed leaves
\citep{malistov2019gbt,marz2024hypertrees}. The model therefore
routes the out-of-support input to its largest leaf. The result is a
$30$--$50\times$ under-prediction of the total before any
correction.\footnote{In error metrics: over $90\%$ mean absolute
percentage error (MAPE), seasonal mean absolute scaled error (MASE)
${\approx}9.3$, and root mean squared scaled error (RMSSE)
${\approx}7.8$. That is about nine times the error of a
seasonal-naive forecast. We report MAPE first because a $>90\%$
aggregate MAPE maps directly to the order-of-magnitude
under-forecast, and we confirm every principal result under the
scale-free metrics MASE and RMSSE
\citep{hyndman2006mase,makridakis2022m5}. MAPE penalizes
over-forecasts more heavily than under-forecasts
\citep{hyndman2006mase}. The collapse is a pure under-prediction, so
MAPE if anything understates it.} On the public M5 data the same
design under-predicts the monthly grand total by $496\times$.

That trees extrapolate poorly is known. What is undocumented is this
specific instance: a global tree collapsing at the hierarchical
aggregate because it is served out of its training support. We
searched for it---across the forecasting journals, the M5 record,
and the gradient-boosting literature---and did not find it.
Section~\ref{sec:related}
explains why the field's standard pipelines never had occasion to
produce it. The nearest published results concern a different
mechanism: error accumulation when bottom-level forecasts are
\emph{summed} \citep{mafildes2022global}. The rest avoid the
operation entirely---the published M5 prize-winning pipelines
produced bottom-level forecasts and summed them; none reports
scoring a tree at the aggregate
\citep{makridakis2022m5,anderer2022topdown}.

\paragraph{Contributions}
\begin{enumerate}[label=\textbf{(C\arabic*)},leftmargin=*]
\item \textbf{The failure, characterized} (Sections~\ref{sec:collapse},
  \ref{sec:public}). The collapse is a training-support failure with a
  simple bound: a tree cannot forecast above the largest target it
  grew, a ceiling the fitted ensembles measurably inherit. It
  reproduces on five panels---at aggregate-to-largest-series
  gaps of $8\times$ to $113\times$, and aggregate-to-median gaps of up
  to $2{,}062\times$---and on three tree libraries. It is invariant
  across eight training seeds. The collapse-versus-fix gap is statistically
  significant on every panel with rolling origins.
\item \textbf{The cure boundary map} (Sections~\ref{sec:cures},
  \ref{sec:transfer}). Three cures work, and all three work the same
  way: they return the target to the model's support. The map has a
  second axis that a one-step accuracy comparison misses. Under
  recursive multi-step forecasting the single-model aggregate-row
  cure re-collapses, a ratio-form target diverges, and per-series
  scaling roughly doubles its error yet stays low. Only seasonal
  differencing keeps its one-step accuracy unchanged, with the
  lowest seed variance. An ablation isolates the load-bearing components of the
  single-model route. A controlled experiment shows that an apparent
  library boundary is a deployment artifact.
\item \textbf{The robustness wall} (Section~\ref{sec:robust}). No
  standard configuration change substitutes for restoring scale. Not
  a longer training window, the series identifier, deeper trees, or
  richer features. Not the Tweedie objective, direct per-horizon
  models, training on the largest series only, or a self-contained
  post-hoc multiplier. Not even pooling every hierarchy level into
  training, which cures a small public hierarchy but is washed out
  at M5's panel size. The one stock flag that abandons constant
  leaves---LightGBM's linear-tree mode---mitigates partially and
  inconsistently, which confirms the mechanism rather than escaping
  it. This negative result consolidates what exists
  in the literature only as scattered single-knob folklore.
\end{enumerate}

\paragraph{What we do not claim} We do not claim the cure.
Per-series scaling is \citet{montero2021globality}'s recommendation,
and it beats our own single-model alternative on one-step error. We
do not claim a corrected tree is the most accurate aggregate
forecaster. The corrected trees land near the seasonal-naive level,
and preventing the collapse, not winning at the aggregate, is the
goal. A practitioner runs a global tree for its value on the
individual series. The aggregate is a coherence constraint the same
model must satisfy when its output feeds planning. A collapse there
blocks deployment no matter how accurate the model is elsewhere.
That is why the failure deserves one paper of careful measurement.

\section{Related work}
\label{sec:related}

\paragraph{The cure is established; the failure is not}
\citet{montero2021globality} apply per-series scaling as routine
preprocessing (their Section~4.3) and measure its effect (their
Section~5.6). Normalization helps, modestly, on their pooled
benchmark sets. Nothing in their experiments involves a hierarchy or
an aggregate. No collapse appears, because their test series lie
inside the pooled training support. The cure is also folk knowledge
in the tooling. The \code{mlforecast} documentation motivates target
transforms because the package ``uses a single global model.''
\code{skforecast} ships differencing as a built-in argument. DeepAR
builds per-series mean scaling into the architecture
\citep{salinas2020deepar}. The cure is everywhere. The failure it
prevents, measured at a hierarchical aggregate, is in none of these
sources.

\paragraph{The favorable evidence, and the apparent counterexample}
\citet{zhao2024localglobal} report a global LightGBM accurate across
hierarchy levels. That appears to contradict our claim. The
resolution is the training support. Their design pools $3{,}049$
small per-product hierarchies of $14$ series each: one product
total, three states, ten stores. The pooled training matrix
therefore contains every aggregate-level series, and each hierarchy
spans only about one order of magnitude. Their model predicts
aggregates it trained on. They credit its top-level accuracy to
sensitivity ``to relatively large values'' seen in training. That is
exactly the configuration our claim excludes. The collapse requires
the aggregate to lie \emph{outside} the training support. Their
result and ours are two sides of the same mechanism.
Section~\ref{sec:robust} tests this resolution by intervention:
pooling every level into training cures a small hierarchy's total,
and fails at M5's panel size, where the in-support rows are washed
out.

\paragraph{Aggregate-level trouble in the M5 record}
The M5 record contains clear signs of the problem, and no
measurement of it. \citet{mafildes2022global} show that the
top-ranked global bottom-up LightGBM pipelines are fragile at the
aggregate levels across test periods. Their mechanism is different
from ours. In their setting the aggregate forecast is the \emph{sum}
of bottom-level forecasts. The fragility comes from cross-series
error covariance---errors that rise and fall together across
series---that the bottom-level loss never sees. As
reported, no model in their study is served out of support. Their most stable pipeline,
GMO, is in fact a LightGBM trained directly \emph{on} the seventy
store-department aggregate series. It is in support at the
aggregate, and stable there---exactly what our mechanism predicts.
The second- and fifth-placed M5 entries adjusted bottom-trained
LightGBM forecasts with multipliers anchored to external top-level
models \citep{makridakis2022m5,anderer2022topdown}. That is
practitioner awareness that bottom-trained trees miss the aggregate,
handled by a workaround rather than diagnosed.
\citet{sprangers2024sparse} improve the aggregate accuracy of
bottom-level trees with a purpose-built sparse hierarchical loss.
That is an intervention beyond any configuration knob, and it is
consistent with our finding that no stock knob suffices. None of
these works serves a tree at the aggregate. The M5 results paper
itself notes that the winners' advantage over benchmarks was largest
at the top level \citep{makridakis2022m5}---because every
competitive pipeline kept the aggregate in support.

\paragraph{Extrapolation limits of trees, and their remedies}
The extrapolation limit itself is documented. Remedies include
leaf-level linear extrapolation \citep{malistov2019gbt}, tree
architectures that learn the parameters of an extrapolating model
\citep{marz2024hypertrees}, and smoothing-anchored calibration in
the M5 setting \citep{lainder2022gbt}. These target within-series
extrapolation along a trend. Our failure is cross-scale: the target
sits at a different order of magnitude than any training series.
The received wisdom, meanwhile, is that scaling does not matter for
trees: ``feature scaling does not noticeably influence their
learning dynamics'' \citep{januschowski2022trees}. That statement
is correct, and it is about inputs at training time. The
collapse is about the target at serving time. A tree is invariant
to monotone (order-preserving) feature rescaling but bounded by its
training targets.
The two properties part ways exactly at an out-of-support
aggregate. Studies that race target transforms measure accuracy,
not stability under recursion
\citep{bansal2026delta,malla2026stationarity}. None measures which
transforms diverge when rolled forward. Section~\ref{sec:recursion}
adds that measurement.

\paragraph{Why the collapse went unreported}
The failure is real on public data---it reproduces on M5
(Section~\ref{sec:public})---yet the hierarchical-forecasting and
M-competition literatures have not reported it. The reason is that
their standard pipelines structurally avoid the operation that
triggers it. Bottom-up pipelines forecast only the bottom level and
sum upward. Optimal-reconciliation methods---OLS combination
\citep{hyndman2011opt} and MinT
\citep{wickramasuriya2019mint}---generate base forecasts at every
level and project them onto the coherent subspace (the set of
forecasts whose levels add up). The aggregate's
base forecast comes from a model fit to the aggregate series
itself, in support. It never comes from a per-series model
extrapolated to aggregate scale. Competition pipelines routinely
scale each series before training. Every one of these choices keeps
the model in support. The collapse therefore requires a specific
configuration: one global model, serving every aggregation level,
on raw unscaled features. That is a production convenience---one
model, one pipeline---rather than a competition design. Our
production setting forces exactly this configuration. A companion
evaluation study on the same class of panels excludes panel-trained
models from its aggregate comparison for precisely this reason. The
contribution is naming and measuring the failure where it is
consequential, not rediscovering that trees extrapolate poorly.

\section{Data and setting}
\label{sec:data}

The production panel is anonymized monthly GTV per buyer from the
transaction warehouse of an on-demand business-to-business
field-service marketplace. The marketplace pairs several thousand
enterprise service-buyer firms with tens of thousands of independent
providers. The panel covers Jan~2021--Dec~2025 ($T{=}60$ training
months; $5{,}658$ buyer series). Per-buyer monthly GTV is
intermittent and heavy-tailed, because demand arrives as discrete,
project-shaped enterprise commitments. The marketplace aggregate is
comparatively smooth, because buyer-specific noise partially cancels
in the sum. All monetary quantities are reported in scale-free form:
ratios, or values indexed to a base period. Absolute GTV levels are
withheld for commercial confidentiality. The analysis is invariant
to this scaling, and the collapse is independently reproduced on
public data (Section~\ref{sec:public}). The training cutoff is the
latest complete warehouse month. Evaluation uses a strict five-month
holdout, Jan--May~2026. No holdout actuals inform selection,
hyperparameters, or fitting. The \emph{aggregate} target is the full
marketplace total---the sum over all active buyers---not a cohort
subtotal.

\begin{figure}[t]
\centering
\includegraphics[width=0.95\linewidth]{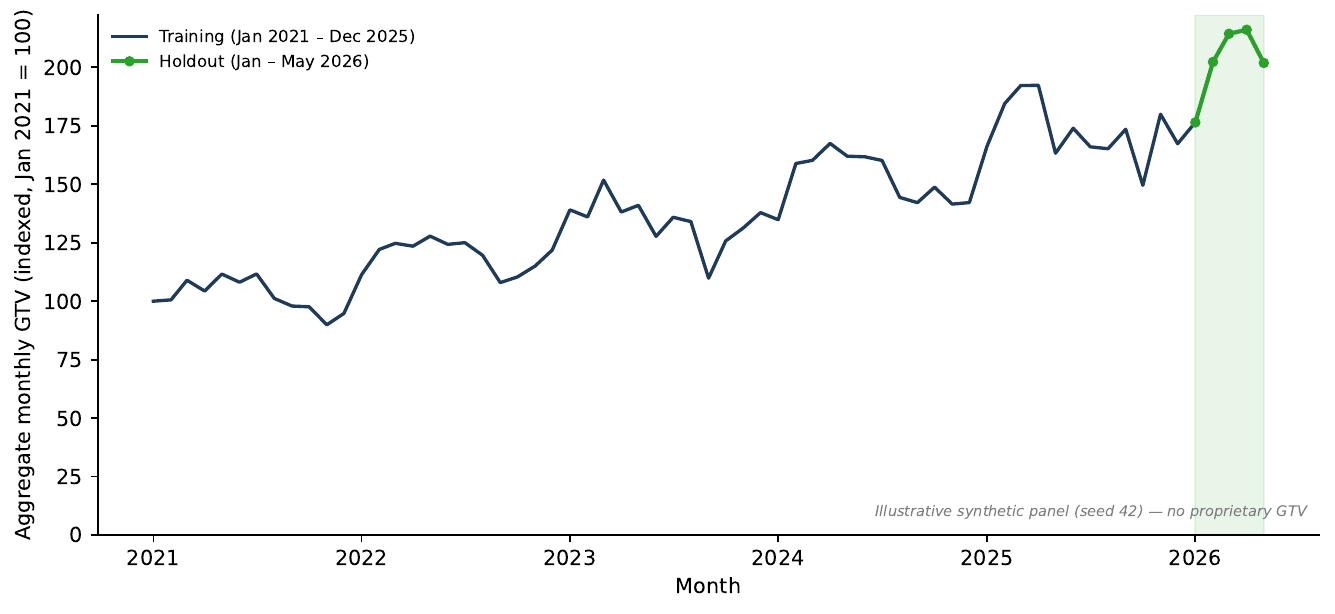}
\caption{Full marketplace aggregate monthly GTV (summed over all
buyers, indexed to Jan~2021${=}100$), Jan~2021--May~2026, shown on a
calibrated synthetic surrogate (Section~\ref{sec:data}; not the
proprietary series). The aggregate trends upward with a recurring
seasonal cycle; every calendar year posts higher GTV than the prior.
The five-month held-out evaluation window (Jan--May~2026) is shaded.
The per-buyer series feeding this total are lumpy, intermittent,
heavy-tailed, and short---the gap this paper turns on.}
\label{fig:overview}
\end{figure}

\paragraph{Structural panel statistics (load-bearing)}
Three properties of the raw panel matter for what follows.
(i)~About $73.5\%$ of buyer-month cells are zero or absent.
(ii)~The median active-month count per buyer is seven months.
(iii)~The aggregate is seasonal on an upward trend across the
window. The combination---
sparse, short, seasonal with an upward trend---is the profile to
which this paper's conclusions transfer. To avoid disclosing
commercially sensitive level information, Figure~\ref{fig:overview}
illustrates this profile on a seeded synthetic surrogate, not the
proprietary series. The surrogate is calibrated to the three
statistics above ($73.8\%$ cell sparsity, seven-month median active
history, a seasonal aggregate on a rising trend). It ships in the
replication package, so the orientation figure is fully
reproducible. The real-panel statistics reported here remain the
load-bearing quantities.

\begin{figure}[tbp]
\centering
\includegraphics[width=0.95\linewidth]{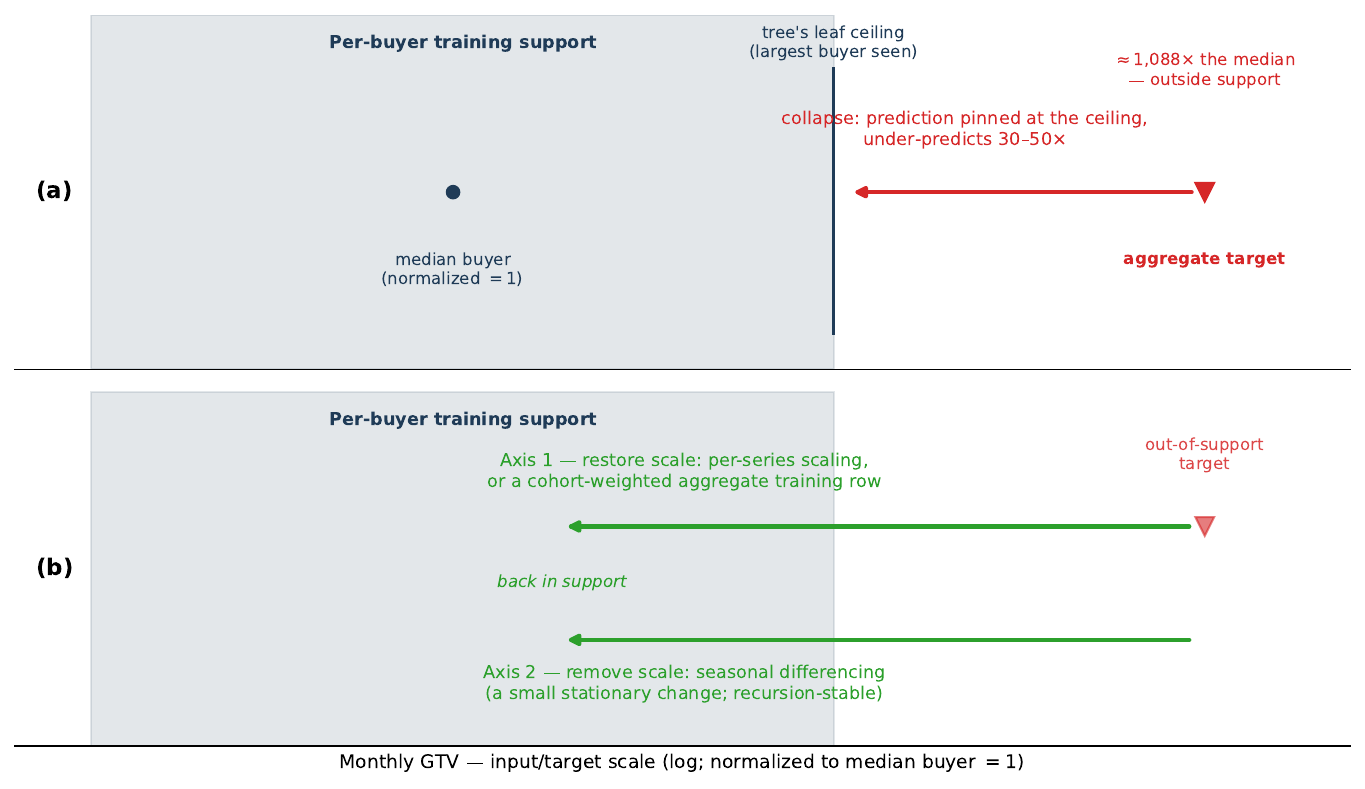}
\caption{Why a global tree collapses at the aggregate, and the two
cures. \textbf{(a)} A regression tree extrapolates as a constant: it
cannot emit a value above its largest leaf---the largest buyer seen
in training (the ceiling). The marketplace aggregate sits
$\approx 1{,}088\times$ above the median buyer and outside the
per-buyer training support, so the out-of-support input is routed to
the ceiling and the forecast under-predicts the total by
$30$--$50\times$. \textbf{(b)} Either axis returns the target to
within support: restoring scale---per-series scaling, or a single
cohort-weighted aggregate training row---or removing
scale---seasonal differencing, the route least harmed by recursive
multi-step forecasting. Conceptual schematic; the
quantities are those reported in this section.}
\label{fig:mechanism}
\end{figure}

\paragraph{The scale gap (the proximate cause)}
The aggregate-series Jan-2026 lag-1 input is about $1{,}088\times$
the median per-buyer lag-1 input. It lies entirely outside the
per-buyer training support. This is the quantity that breaks an
absolute-lag global tree. Figure~\ref{fig:mechanism} summarizes the
mechanism and the two axes that return the target to support.

\section{Experimental protocol}
\label{sec:method}

\paragraph{The design under test} We test the production convenience
configuration of Section~\ref{sec:related}: one global
gradient-boosted tree, trained on all individual series of a panel
with absolute lag features, then asked to forecast the hierarchical
aggregate directly. Trees are fit on each panel's full pinned
training window: $60$ months on the production panel, $48$ on the
synthetic panel, $52$ on M5, and $216$ on Tourism at the main
origin. A window sweep from $36$ to $120$ months finds the collapse
at every length (Section~\ref{sec:robust}).
Forecasts are either one-step---teacher-forced, every lag input a
realized value---or recursive multi-step, each step fed back as the
next step's lag input. We report both throughout, because the two regimes
separate the cures (Section~\ref{sec:cures}). Seed counts vary by
experiment family, and each table states its own. The synthetic
ablation uses six data seeds, the cures comparison five, the
seed-stability band eight, and the linear-leaf probes and the
onset sweep three each. Each experiment holds
the panel, features, window, and holdout fixed. It varies exactly
one thing: the cure, the component, the library, or the
configuration knob under test.

\paragraph{Pinned configuration and provenance}
Every result artifact was produced against one configuration on a
short, strictly additive commit history. The configuration: input
vintage \code{gtv\_by\_month\_by\_buyer\_20260614}, training cutoff
Dec~2025, holdout Jan--May~2026, and the fix recipe. Each artifact
records its commit SHA, a clean-tree flag, the seed, and library
versions in a provenance sidecar (Section~\ref{sec:repro}). We make
the configuration explicit for a reason. An earlier preprint
version prominently reported a $5.4\%$ aggregate MAPE for the corrected
LightGBM. That number depended on a regime-step exogenous indicator
since removed from the shipped code. It is not reproducible from
current code, and we do not rely on it. It is distinct from the
per-series-scaling one-step MAPE of $5.4\%$ in
Table~\ref{tab:scaling}, a current result from a different recipe.

\paragraph{Metrics and significance}
We report MAPE at the aggregate. For the collapse-and-fix result we
also report the scaled metrics seasonal MASE \citep{hyndman2006mase}
and RMSSE, evaluated across rolling origins. The central finding
therefore does not rest on a single metric or split. The scaled
metrics divide by an in-sample naive mean absolute error whose
season $m$ we set by estimability,
\begin{equation}
d_m(z)=\frac{1}{n-m}\sum_{s=m+1}^{n}\bigl|z_s-z_{s-m}\bigr|,
\label{eq:mase-denom}
\end{equation}
with $m{=}12$ at the long, seasonal aggregates. On the short public
per-series panels a seasonal denominator is not estimable, and
$m{=}1$ is used. The collapse-versus-fix gap is tested with the
Diebold--Mariano test \citep{diebold1995} under the
Harvey--Leybourne--Newbold small-sample correction
\citep{harvey1997dm}. The test runs on the per-step
absolute-percentage-error loss differential pooled across rolling
origins. A Wilcoxon signed-rank test \citep{wilcoxon1945} across
origins cross-checks it (Section~\ref{sec:significance}). Stochastic
methods are reported as mean$\pm$sd bands across seeds
(Section~\ref{sec:seed}); deterministic methods carry zero seed
variance.

\section{The collapse}
\label{sec:collapse}

\begin{figure}[t]
\centering
\includegraphics[width=0.9\linewidth]{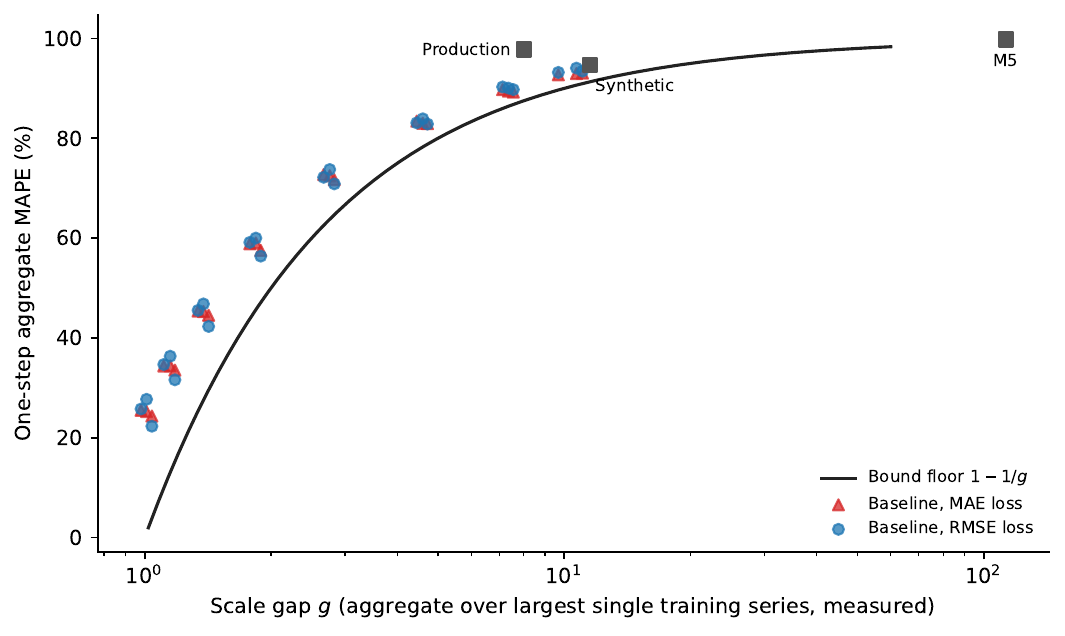}
\caption{The onset of the collapse tracks the support bound.
One-step aggregate MAPE of the baseline global tree against the
measured scale gap $g$ (aggregate over largest single training
series), swept on the synthetic panel via a dominant series that
holds shape fixed while scale varies; three data seeds and both
losses per gap setting. Every point sits at or above the $1-1/g$ floor
(solid line), the error rises monotonically with the gap, and the
grey squares anchor the sweep to the paper's measured panels. A
$1.15\times$ gap already produces $34\%$ error: there is no safe
band above $g{=}1$.}
\label{fig:onset}
\end{figure}

\paragraph{The mechanism, precisely} A single regression tree is
piecewise-constant. Each prediction is one leaf value, and a leaf
value is an average of training targets. A single tree is therefore
bounded by its training range,
\begin{equation}
\min_{i,t} y_{i,t}\;\le\; f(x)\;\le\;\max_{i,t} y_{i,t}
\qquad\text{for every input }x,
\label{eq:leafbound}
\end{equation}
with $i$ indexing series and $t$ months. A boosted forecaster is a
sum of many trees plus a base value, and no exact bound of this
form caps the sum. In practice the cap holds anyway. An input far
above the training range falls into every tree's outermost leaf, so
the ensemble emits close to the largest value it fitted. We verify
this directly. Across the $48$ fitted baselines of the onset sweep
below (Figure~\ref{fig:onset}), no emission at the aggregate
exceeded the largest training target; the largest measured ratio
was $0.95$. The training maximum is thus an empirical ceiling for
the boosted model, and an exact bound only for a single tree fit
directly to the targets. Write
the aggregate level as $A=\sum_i y_{i,t}$. Write the ceiling as
$c=\max_{i,t} y_{i,t}$, the largest single-series level the model
saw. An aggregate query is routed to near-ceiling leaves, so
$\hat A\le c$ in every fit we measure. The error is then bounded
below by a quantity that depends only on the \emph{scale gap}
$g\equiv A/c$:
\begin{equation}
\operatorname{MAPE}=1-\frac{\hat A}{A}\;\ge\;1-\frac{c}{A}=1-\frac1g .
\label{eq:collapse}
\end{equation}
The collapse is a structural ceiling, not a tuning failure. On the
production panel the aggregate sits about $8\times$ above the
largest single training series, so the floor is already $87.5\%$.
The realized ceiling sits lower still, because a leaf value is an
\emph{average} of training targets, not a maximum. We measure each
library's realized ceiling directly: the largest value the fitted
baseline model emits when queried at the aggregate. The measured
aggregate-to-ceiling gaps are $42.7\times$ for LightGBM and
$51.9\times$ for CatBoost, which matches the observed
$30$--$50\times$ under-prediction. The same measurement on the
public panels closes the loop (Table~\ref{tab:ceilings}). On every
public panel and under both losses, the ceiling-implied floor sits
no more than $3.4$ points below the measured collapse; the measured
aggregate-to-ceiling gaps run $11.6\times$ to $488\times$. No knob that keeps
piecewise-constant leaves lifted a measured ceiling above the
training range (Section~\ref{sec:robust}). This ceiling gap is
distinct from, and smaller than, the $\approx1{,}088\times$
aggregate-to-median-buyer gap that puts the input out of support in
the first place. Every cure in Section~\ref{sec:cures} works by
shrinking $g$ to $O(1)$: raise $c$ to aggregate scale, rescale $A$
down to per-series scale, or remove the level entirely.

\paragraph{The onset tracks the bound} The five panels of this
paper all sit deep in the collapse regime ($g\ge8$, floor
$\ge87.5\%$). To map the transition below that, we sweep the gap
through $g\approx1$--$10$ on the synthetic panel. One dominant
series carries the aggregate's own shape, so the sweep changes
scale and nothing else (three data seeds, both losses, $48$ fits).
Figure~\ref{fig:onset} shows the result. The measured one-step
aggregate MAPE exceeds the $1-1/g$ floor on every run and rises
monotonically with the measured gap (Spearman rank correlation
$+1.0$ under both losses). The damage starts immediately. A gap of just $1.15\times$
already produces $34\%$ aggregate MAPE, and $2\times$ produces
$59\%$. There is no safe band above $g=1$, and the bound is not
just a floor: as the gap grows, the measured error converges to it.

\begin{table}[t]
\centering\small
\caption{The measured ceilings close the loop on every public
panel. For each fitted baseline, the realized ceiling $\hat c$ is
the largest value the model emits when queried at the aggregate;
the implied floor is $1-\hat c/A$, the bound of
Eq.~\eqref{eq:collapse} at the measured gap. Under both losses the
floor sits no more than $3.4$ points below the measured one-step
collapse: the support bound does not just permit the collapse, it
predicts its size. One fit per cell (tree seed $42$); no emission
exceeded the largest training target (largest ratio $0.762$).}
\label{tab:ceilings}
\begin{tabular}{llrrr}
\toprule
Panel & Loss & Gap $A/\hat c$ & Implied floor & Measured MAPE \\
\midrule
Synthetic & MAE & $14.5\times$ & $93.0\%$ & $93.3\%$ \\
Synthetic & RMSE & $11.6\times$ & $91.3\%$ & $92.2\%$ \\
Australian Tourism & MAE & $17.2\times$ & $94.0\%$ & $96.2\%$ \\
Australian Tourism & RMSE & $14.5\times$ & $92.9\%$ & $96.3\%$ \\
M5 & MAE & $487.9\times$ & $99.8\%$ & $99.8\%$ \\
M5 & RMSE & $271.6\times$ & $99.6\%$ & $99.7\%$ \\
UCI Online Retail~II & MAE & $45.5\times$ & $97.6\%$ & $97.6\%$ \\
UCI Online Retail~II & RMSE & $22.0\times$ & $95.1\%$ & $96.1\%$ \\
\bottomrule
\end{tabular}
\end{table}

The unmodified design under-predicts the aggregate by
$30$--$50\times$ on three independent libraries. The libraries share
feature engineering but no code. The failure is therefore a property
of the design pattern, not of any one library. DeepAR, a global
model with built-in per-series mean scaling
\citep{salinas2020deepar}, escapes the catastrophic collapse. Its
median aggregate MAPE is $12.5\%$ across five seeds (range
$6$--$15\%$), far from the trees' $30$--$50\times$
under-prediction, though still not competitive (MASE $1.3$, just
above a seasonal-naive forecast). This is consistent with the
absolute-scale mechanism rather than with global pooling as the
root cause. An ablation isolates the variable. On the synthetic
panel we train the same DeepAR twice, changing only the
mean-scaling flag, three seeds each. With scaling on, the
aggregate forecast is accurate ($3.9$--$9.0\%$ MAPE). With scaling
off, it collapses completely: the network emits near zero at the
aggregate query, $100\%$ MAPE on every seed. The failure signature
differs from the trees'---an emission near zero rather than near
the ceiling---but the enabling condition, a query far outside the
training scale, is the same. Scaling, not the architecture, is the
escape.

\section{The cures, and where each one fails}
\label{sec:cures}

Three cures work. All three return the target to the model's
support. We present the two scale cures first, then the trend-axis
cure, and then the boundary that separates all three: behavior
under recursive multi-step forecasting.

\subsection{Per-series scaling (known)}
\label{sec:scaling}

The standard way global models handle disparate scales is
per-series scaling: divide each series by its own level before
fitting, multiply back afterwards. \citet{montero2021globality}
recommend it, and DeepAR does it by construction
\citep{salinas2020deepar}. We apply it to the same global tree.
Each series is divided by one plus its training mean; the aggregate
is scaled by its own mean at inference:
\begin{equation}
\tilde y^{(i)}_t=\frac{y^{(i)}_t}{1+\bar y^{(i)}},\quad
\bar y^{(i)}=\frac{1}{n_i}\sum_t y^{(i)}_t,\qquad
\hat y^{(i)}_{t+h}=\bigl(1+\bar y^{(i)}\bigr)\,\hat{\tilde y}^{(i)}_{t+h}.
\label{eq:scaling}
\end{equation}
This maps every series, the aggregate included, to $O(1)$. It
restores the aggregate to the model's support and prevents the
collapse on every panel (Table~\ref{tab:scaling}). It is the most
accurate one-step cure on two of the four panels (production
$5.4\%$; M5 $8.2\%$, five-seed means) and the second most accurate
on the other two, behind seasonal differencing. It is also
library-agnostic, and it requires no seasonal lag. We recommend it
as the default cure for one-step forecasts. We repeat
that it is not ours. The finding here is that the recommended
preprocessing step is \emph{load-bearing}: omit it in this
configuration and the forecast does not degrade, it collapses.

\subsection{A single-model alternative: the cohort-weighted aggregate-level training row}
\label{sec:fix}

Some deployments keep one un-normalized global model spanning every
aggregation level, with no per-series scale bookkeeping. Our
production system trains this configuration. At serving time it
forecasts the aggregate with in-support methods; the tree serves
only the individual series. For that configuration there is a
single-model route: insert the cross-buyer total into training as
one extra series, so the tree observes the high-lag region. The
recipe has four parts. (a)~A $\log(1{+}y)$ transform of target and
lags compresses the scale gap into a contiguous range.
(b)~Squared-error (RMSE) loss replaces MAE, because under MAE on
intermittent data the trees collapse to a near-zero constant leaf.
(c)~The aggregate-level training row is weighted by
$w_{\mathrm{agg}}=n_{\mathrm{buyer}}/n_{\mathrm{agg}}$, so its rows
reach parity with the $\approx 211{,}000$ per-buyer rows. (The row
count sits below buyers${}\times{}$months because a buyer-month
becomes a training row only when its full lag and rolling feature
history, including the $12$-month lag, exists.) (d)~A
recursive-step guardrail caps each predicted step at $2\times$ the
trailing-12-month buffer maximum.

\begin{table}[t]
\centering\small
\caption{Cumulative ablation of the four-component recipe on a
synthetic hierarchical panel (one-step aggregate MAPE, mean over six
data seeds; Section~\ref{sec:synthetic}). Component (c)---the
cohort-weighted aggregate-level row---is the load-bearing lever; (a)
and (b) do nothing alone because the collapse is a training-support
problem, not a loss or transform problem.}
\label{tab:ablation}
\begin{tabular}{lr}
\toprule
Configuration & One-step aggregate MAPE \\
\midrule
Baseline: absolute lags, MAE, per-buyer only & 94.7\% \\
$+$ (a) $\log(1{+}y)$ target/lags & 94.5\% \\
$+$ (b) RMSE loss on log scale & 95.4\% \\
$+$ (c) cohort-weighted aggregate row & \textbf{17.5\%} \\
$+$ (d) $2\times$ recursive guardrail (full fix) & \textbf{17.5\%} \\
\bottomrule
\end{tabular}
\end{table}

The cumulative ablation (Table~\ref{tab:ablation}) shows the
construction path. Recovery happens entirely when the
cohort-weighted row~(c) is added ($95\%\to17.5\%$). A cumulative
table cannot establish necessity. We therefore also run a
leave-one-out ablation---start from the full recipe, remove one
component at a time---on both the synthetic and the production panel
(Table~\ref{tab:loo}). Two components are load-bearing, and they
interact. The aggregate row~(c) supplies the only aggregate-scale
training signal; removing it reverts the collapse on both panels.
The RMSE loss~(b) is what lets that single high-magnitude row move
the global fit. Under MAE the lone aggregate row is washed out among
${\approx}10^5$ per-buyer rows, so removing~(b) while keeping the
row also reverts the collapse. Weighting the row is decisive at
production's extreme scale gap: removing the weight sends
$15\%\to80\%$. The log transform~(a) and the guardrail~(d) are not
individually necessary in either panel. We retain them as defensive
refinements. The intended value of this route over per-series scaling is
cold start. A new series has no history from which to estimate a
scale, so a scale-based cure has nothing to divide by, while this
route keeps no per-series bookkeeping. We state that as a design
rationale: this paper does not measure cold-start accuracy.

\begin{table}[t]
\centering\small
\caption{Leave-one-out ablation: one-step aggregate MAPE when each
component is removed from the full recipe, on the synthetic panel
(mean over six seeds) and the production panel (pinned
Jan--May~2026). Two components are load-bearing on both panels and
interact; the log transform and the recursive cap are defensive
refinements, not individually necessary.}
\label{tab:loo}
\footnotesize
\setlength{\tabcolsep}{4pt}
\begin{tabular}{lrrl}
\toprule
Configuration & Synthetic & Production & Role \\
\midrule
Full fix (reference) & 17.5\% & 15.1\% & --- \\
$-$ (c) cohort-weighted aggregate row & 95.4\% & 97.8\% & load-bearing \\
$-$ (b) RMSE loss (use MAE) & 202\% & 99.6\% & load-bearing \\
$\;\;-$ cohort weight only (row kept) & 31.1\% & 79.6\% & decisive at extreme gap \\
$-$ (a) $\log(1{+}y)$ transform & 15.7\% & 18.6\% & refinement \\
$-$ (d) $2\times$ recursive guardrail & 17.5\% & 15.1\% & dormant guardrail \\
\bottomrule
\end{tabular}
\end{table}

\subsection{A second axis: a stationary target, and the recursion boundary}
\label{sec:recursion}

Restoring scale is not the only way back into support. A
complementary axis treats the target instead. Predict a seasonal
difference $y_t-y_{t-12}$ rather than the level, and reconstruct
each step by adding back the realized seasonal lag:
\begin{equation}
d_t=y_t-y_{t-12},\qquad \hat y_{t+h}=y_{t+h-12}+\hat d_{t+h}.
\label{eq:seasdiff}
\end{equation}
A seasonal difference is scale-free. A marketplace and a single
buyer each growing $8\%$ year over year present the same target. The
aggregate is then no longer out of support, and differencing cures
the one-step collapse on every panel (Table~\ref{tab:scaling}).

\begin{figure}[t]
\centering
\includegraphics[width=0.98\linewidth]{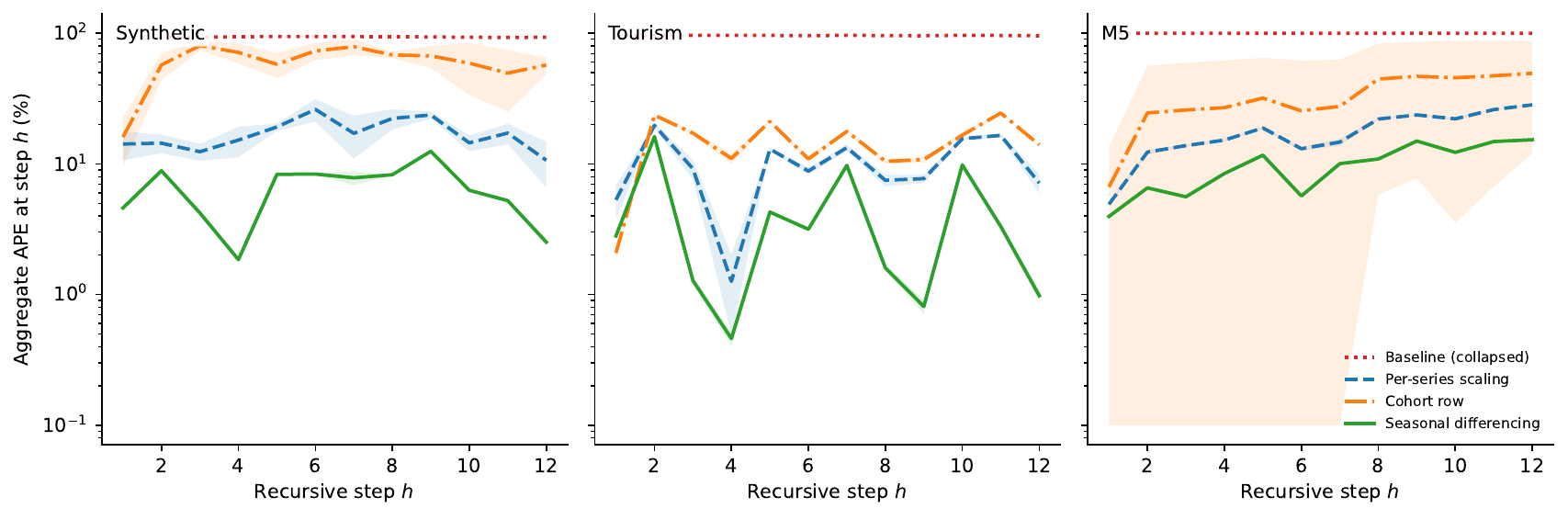}
\caption{Recursive aggregate error, resolved by step (five-seed
mean; shaded bands one standard deviation; log scale). The cohort
row re-collapses immediately on the synthetic panel and climbs a
staircase on M5; per-series scaling drifts on M5; seasonal
differencing is the lowest-error cure at nearly every step. The
window means of Table~\ref{tab:scaling} conceal these dynamics.}
\label{fig:steps}
\end{figure}

Recursion separates the three cures, and the separation is the part
of the cure map we found documented nowhere (Table~\ref{tab:scaling}).
The single-model cohort row is accurate one step
ahead. Rolled forward, it re-collapses (production $15\%\to27\%$,
M5 $15\%\to34\%$, synthetic $15\%\to61\%$ recursive mean), because
each step still asks a
leaf to emit a larger absolute level than any it grew. Per-series
scaling does not re-collapse. It roughly doubles its error at worst
yet stays low (production $5.4\%\to8.2\%$, M5 $8.2\%\to17.9\%$),
and on the production panel it remains the most accurate recursive
cure. Seasonal differencing has the property neither scale cure
offers. It reconstructs each step from the realized seasonal lag
and inherits no leaf ceiling, so its one-step and recursive window
means coincide (production $11.0\%/10.9\%$, Tourism
$4.6\%/4.5\%$; Figure~\ref{fig:steps} resolves the steps). Its
seed variance is the lowest of the cures. The cohort row's
recursive error is seed-fragile (M5 $\pm30.8$). One further
caution. A purely multiplicative target---the ratio
$y_t/y_{t-12}$---compounds under recursion and is unstable on
heavy-tailed panels; its recursive MAPE diverges on the production
panel. The additive seasonal difference, not the ratio, is the
stable form. We recommend per-series scaling for one-step use.
Where forecasts are rolled forward, seasonal differencing is the
predictable choice---its accuracy does not change between
regimes---while scaled levels degrade gradually. The cohort row is
the workaround for a deployment that must keep one un-normalized
model; its seed fragility (Section~\ref{sec:seed}) is the price.

\paragraph{The boundary, step by step} Window means understate how
sharply recursion separates the cures. Figure~\ref{fig:steps}
resolves the recursive error by step. The cohort row's re-collapse
is immediate, not gradual. On the synthetic panel a single step of
fed-back forecasts lifts its error from $16\%$ to $57\%$, and two
more to $80\%$; it then oscillates with the seasonal cycle.
On M5's steep total the same cure climbs a staircase instead, from
$7\%$ at the first step to $49\%$ at the twelfth. Per-series
scaling stays low and trendless on the seasonal panels but drifts
steadily on M5, reaching $28\%$. Seasonal differencing is the
lowest-error cure at nearly every step. Step-resolved, it too
drifts mildly on M5, to $15\%$ at the twelfth step: on a steep
aggregate no cure is exempt from recursion---this one is only the
least harmed.

\begin{table}[t]
\centering\footnotesize
\caption{Three cures across two axes: aggregate MAPE (five-seed
mean$\pm$sd), one-step and recursive, identical recipe. The naive
global tree collapses on every panel. \emph{Scale axis:} per-series
scaling and the single-model cohort-weighted row both restore scale
(scaling the more accurate one-step). \emph{Trend axis:} seasonal
differencing makes the target stationary. Under recursion the
cohort row re-collapses, per-series scaling degrades but stays low,
and seasonal differencing alone keeps its one-step accuracy
unchanged; its seed variance is the
lowest of the cures, while the cohort row's recursive error is
seed-fragile (M5 $\pm30.8$). Values are from the five-seed
cures-comparison run; the synthetic baseline ($93.2\%$) is thus
${\approx}1.5$ points below the six-seed cumulative-ablation
baseline ($94.7\%$, Table~\ref{tab:ablation}), run-to-run variation
immaterial to the collapse.}
\label{tab:scaling}
\begin{tabular}{lrrrr}
\toprule
Dataset & Baseline & Per-series & Cohort row & Seasonal diff. \\
\midrule
\multicolumn{5}{l}{\textit{One-step aggregate MAPE}}\\
Synthetic  & $93.2\pm0.0$ & $11.7\pm0.7$ & $14.9\pm0.9$ & $\mathbf{6.9\pm0.1}$ \\
Production & $97.8\pm0.0$ & $\mathbf{5.4\pm0.3}$ & $15.1\pm0.3$ & $11.0\pm0.0$ \\
M5         & $99.8\pm0.0$ & $\mathbf{8.2\pm0.1}$ & $14.7\pm1.5$ & $10.0\pm0.0$ \\
Tourism    & $96.1\pm0.1$ & $9.6\pm0.6$ & $14.0\pm0.1$ & $\mathbf{4.6\pm0.0}$ \\
\midrule
\multicolumn{5}{l}{\textit{Recursive aggregate MAPE}}\\
Synthetic  & $93.5\pm0.1$ & $17.2\pm0.5$ & $61.4\pm4.5$ & $\mathbf{6.6\pm0.1}$ \\
Production & $98.0\pm0.0$ & $\mathbf{8.2\pm1.3}$ & $27.1\pm10.6$ & $10.9\pm0.0$ \\
M5         & $99.8\pm0.0$ & $17.9\pm0.2$ & $33.6\pm30.8$ & $\mathbf{10.0\pm0.0}$ \\
Tourism    & $96.1\pm0.1$ & $10.4\pm0.1$ & $15.0\pm0.2$ & $\mathbf{4.5\pm0.0}$ \\
\bottomrule
\end{tabular}
\end{table}

\section{The cures transfer across libraries}
\label{sec:transfer}

Given the identical single-model recipe---including the load-bearing
cohort weight---the collapse is prevented for all three major
boosted-tree libraries
\citep{ke2017lightgbm,chen2016xgboost,prokhorenkova2018catboost} on
every panel we test. On the production panel we sweep full-recipe
CatBoost over depth $\in\{4,6,8,10\}$, learning rate
$\in\{0.03,0.05,0.1\}$, and $L_2 \in\{1,3,9\}$: 36 configurations.
Every configuration recovers the aggregate. The range is
$14.9$--$18.5\%$ one-step MAPE (median $15.9\%$), statistically tied
with LightGBM ($15.1\%$) on the identical panel.

\paragraph{An apparent library boundary is a deployment artifact}
The shipped production configuration lists CatBoost at $92.9\%$
aggregate MAPE, collapsed. That result once read as ``CatBoost
cannot forecast the aggregate.'' It is not a library limitation. The
production CatBoost path is trained without the cohort-balanced
sample weight. The leave-one-out (Table~\ref{tab:loo}) shows that
withholding this weight reverts the collapse for any library: it
sends the full-recipe LightGBM from $15\%$ to $80\%$ on the same
panel. Supplied the full recipe, CatBoost recovers. A controlled
isolation confirms the weight is the variable, not the dataset or
the library. CatBoost recovers with the cohort-weighted row and
collapses with an unweighted row, on both M5 ($16.8\%$ vs.\
$94.4\%$) and production ($15.7\%$ vs.\ $87.0\%$). On a synthetic
panel we sweep the scale gap from $38\times$ to $1{,}568\times$
under the full recipe. CatBoost tracks LightGBM throughout, both in
the $10$--$24\%$ band. There is no scale-gap threshold beyond which
CatBoost alone fails. CatBoost is somewhat more fragile under
recursive inference at the extreme production gap: recursive
five-step MAPE spans $18$--$70\%$ across the grid, and most
configurations land above LightGBM's ${\approx}37\%$. That is
tunable, not categorical: at depth~$6$
and learning rate $0.1$ its recursive MAPE is $18\%$, better than
LightGBM. We find no genuine cross-library boundary.

\section{The collapse reproduces on public data}
\label{sec:public}

\subsection{Synthetic reproduction}
\label{sec:synthetic}

\begin{figure}[t]
\centering
\includegraphics[width=0.92\linewidth]{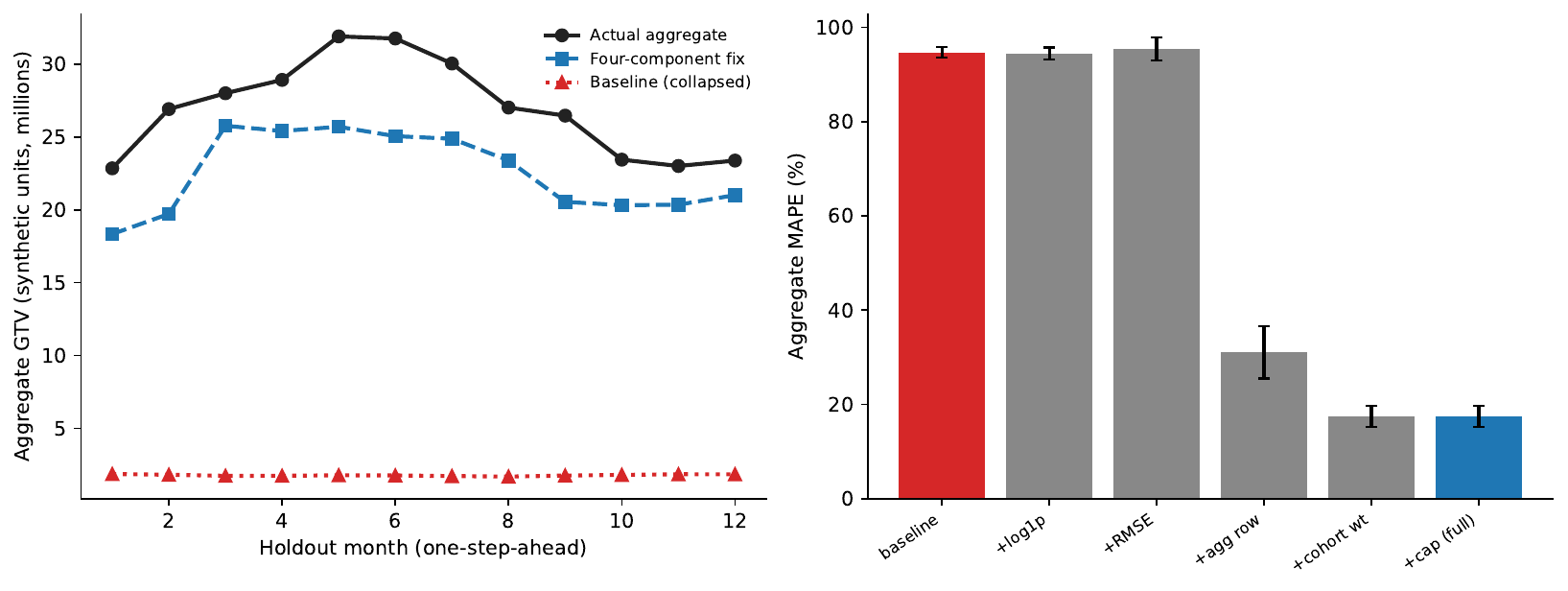}
\caption{Synthetic reproduction of the collapse and fix. The
baseline global tree (flat line) under-predicts the smooth synthetic
aggregate; the cohort-weighted recipe tracks it. Mechanism
confirmation, not MAPE replication---synthetic per-buyer series
carry no learnable cross-buyer structure, so recovery stops at
$17.5\%$.}
\label{fig:synthetic}
\end{figure}

The failure is a property of the design pattern, not of the
proprietary data. To show this we build a synthetic hierarchical
panel (300 buyers $\times$ 60 months) that carries no exploitable
cross-buyer structure. It reproduces only the conditions that drive
the collapse: a scale gap, buyer-specific seasonal phases (so the
aggregate is smooth by cancellation), and about $7\%$ yearly
growth. The synthetic aggregate is $11.5\times$ the largest
single-buyer series---out of per-buyer support. The baseline tree
collapses to $94.7\%$ one-step aggregate MAPE across six data
seeds. The four-component recipe recovers it to $17.5\%$
(Table~\ref{tab:ablation}, Figure~\ref{fig:synthetic}). The
residual $\approx 17\%$ is the same leaf-ceiling mechanism in
milder form: a growing aggregate climbs above its highest training
leaf. That residual is why a deployment should pair a corrected
tree with monitoring rather than trust a bare tree to extrapolate.

\subsection{M5}
\label{sec:m5}

\begin{figure}[t]
\centering
\includegraphics[width=0.92\linewidth]{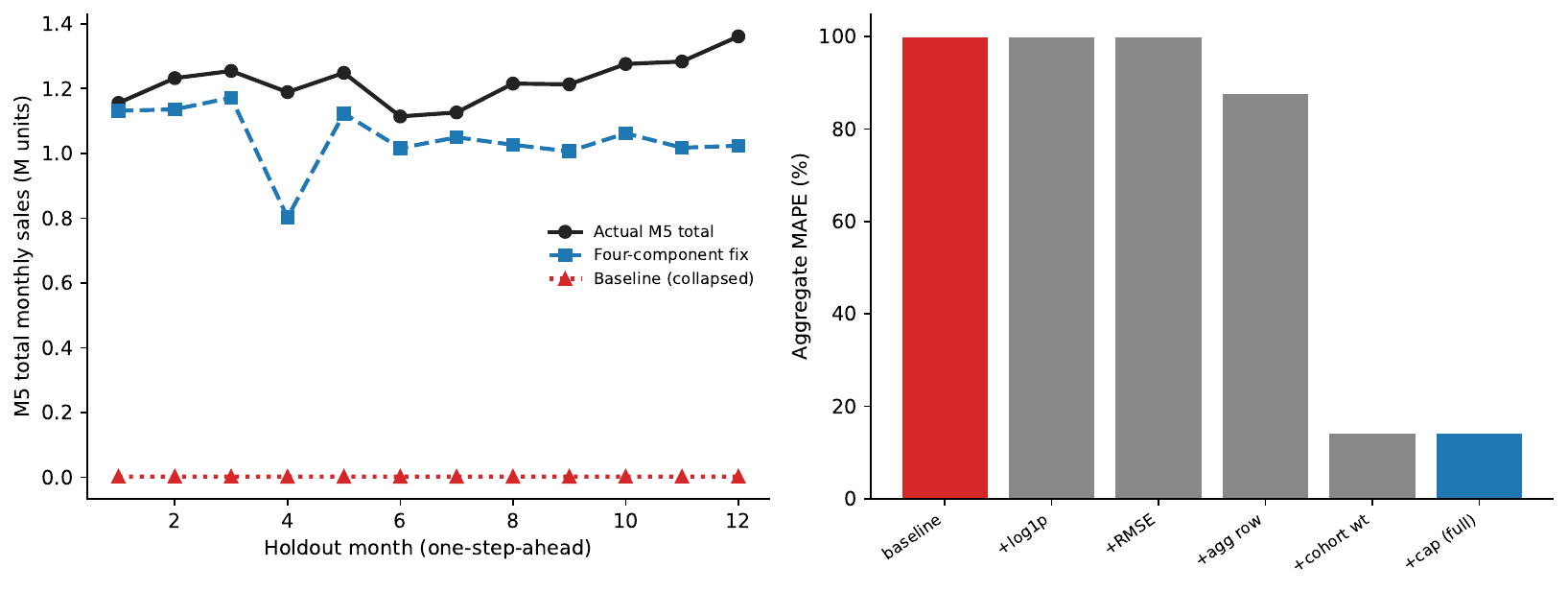}
\caption{Collapse and fix replicated on public M5. Left: the
baseline global tree collapses on the M5 grand total while the
four-component recipe tracks it. Right: the cumulative ablation,
with the cohort-weighted aggregate row as the decisive lever---the
same structure as the synthetic (Figure~\ref{fig:synthetic}) and
production panels.}
\label{fig:m5}
\end{figure}

We replicate the collapse and the recipe on the public M5
competition dataset \citep{makridakis2022m5}, the canonical
hierarchical forecasting benchmark, using identical feature, fix,
and recursive-forecast code. M5's 30{,}490 bottom item-store series
aggregate to one grand total. Aggregated to monthly, the total is
$113\times$ the largest single series in training---the same
out-of-support condition. (M5 is canonically a daily, $28$-day,
RMSSE-scored competition. We re-aggregate to monthly to match the
production cadence and to test the cross-scale mechanism, not
competition-ranking accuracy.) The naive global tree collapses to
$99.8\%$ aggregate MAPE, a $496\times$ under-prediction. The
cumulative ablation reproduces the production structure exactly.
The log transform and RMSE loss do nothing on their own
($99.8\%\to99.8\%$). An unweighted aggregate row drops it to
$87.5\%$. The cohort-weighted row is again the decisive lever
($\to14.1\%$). The recipe recovers LightGBM ($14.1\%$ one-step /
$45.1\%$ recursive twelve-month) and XGBoost ($6.3\%$ / $10.3\%$)
(Table~\ref{tab:m5}, Figure~\ref{fig:m5}). The $45.1\%$ recursive
value is a single pinned run; the five-seed cures-comparison mean
for the same quantity is $33.6\%\pm30.8$
(Table~\ref{tab:scaling})---the cohort row's recursive seed
fragility, documented in Section~\ref{sec:seed}. The $14.1\%$ is the
single $12$-month holdout reported here; the ten-origin
rolling-mean LightGBM fix is $12.6\%$
(Table~\ref{tab:significance}).

\begin{table}[t]
\centering\small
\caption{Public replication on M5 (30{,}490 monthly item-store
series, 12-month holdout). One-step aggregate MAPE. The collapse and
the four-component recipe reproduce; the cohort-weighted aggregate
row is the load-bearing lever, exactly as on the synthetic and
production panels.}
\label{tab:m5}
\begin{tabular}{lrr}
\toprule
Configuration / library & Baseline & Four-component fix \\
\midrule
LightGBM (ablation) & 99.8\% & \textbf{14.1\%} \\
XGBoost & 99.9\% & \textbf{6.3\%} \\
CatBoost & 99.8\% & \textbf{16.8\%} \\
\bottomrule
\end{tabular}
\end{table}

The M5 result is not an artifact of MAPE or of a single split.
Across three rolling origins, under the competition's own scaled
metrics, the baseline collapses on every origin and metric (MASE
$8.5$--$12.4$, roughly ten times worse than a seasonal-naive
forecast). The recipe recovers on every origin (MASE $1.0$--$2.9$).
The fix lands near the seasonal-naive level, not below it. In
scaled-metric terms, the contribution is collapse-prevention, not
an aggregate-accuracy win.

\subsection{Australian Tourism}
\label{sec:tourism}

M5 could be a retail-specific coincidence. To guard against that,
we replicate on a second public hierarchical benchmark from an
unrelated domain: the Australian Tourism dataset of monthly visitor
nights \citep{wickramasuriya2019mint}, the canonical hierarchical
set of the reconciliation literature. Its $304$ bottom
region-by-purpose series sum to a national total about $598\times$
the median series level in training, and about $8\times$ the
largest---still outside every series' support. Using identical
code, the naive global tree collapses ($96.2\%$ one-step aggregate
MAPE, seasonal MASE $16.4$). Recovery again happens entirely at the
aggregate-row step ($95.9\%\to13.5\%$ for the unweighted row in the
cumulative ablation, $14.0\%$ cohort-weighted; log and RMSE alone
do nothing). The full recipe recovers all three
libraries (Table~\ref{tab:tourism}). The collapse-and-fix holds
across three rolling origins (baseline MASE $15.5$--$16.4$ every
origin; fix MASE $1.55$--$2.30$).

\begin{table}[t]
\centering\small
\caption{Second public replication on Australian Tourism ($304$
monthly series, $12$-month holdout, ${\approx}598\times$
aggregate-to-median gap).
One-step aggregate MAPE / seasonal MASE. The collapse and the
library-agnostic recipe reproduce in a non-retail domain.}
\label{tab:tourism}
\begin{tabular}{lrr}
\toprule
Configuration / library & MAPE & MASE \\
\midrule
Baseline (LightGBM, no fix) & 96.2\% & 16.4 \\
Full fix --- LightGBM & \textbf{14.0\%} & 2.30 \\
Full fix --- XGBoost & \textbf{5.2\%} & 0.90 \\
Full fix --- CatBoost & \textbf{11.0\%} & 1.84 \\
\bottomrule
\end{tabular}
\end{table}

\subsection{A public business-buyer panel}
\label{sec:b2b}

The production panel is a field-service marketplace. Is the
collapse sector-specific? We replicate on a public
business-to-business buyer panel from an unrelated industry,
country, and decade: UCI Online Retail~II
\citep{chen2012onlineretail}, a UK online retailer whose customers
are largely wholesale businesses. The panel has $5{,}878$ business
customers, monthly revenue, and a ${\approx}2{,}062\times$
aggregate-to-median scale gap. The baseline collapses ($98.7\%$ one-step aggregate MAPE, MASE $9.5$).
The four-component recipe prevents it ($36.8\%$, MASE $3.1$), with
recovery again landing at the cohort-weighted row. The panel spans
only $25$ months, too short for strong seasonal modeling. The
recipe therefore prevents the collapse (a $2.7\times$ error
reduction) rather than reaching single digits. We report the
one-step scale effect, not seasonal accuracy.
Table~\ref{tab:generalization} collects all five panels, ordered
by scale gap.

\begin{table}[t]
\centering\small
\caption{Generalization of the collapse and the four-component
recipe across all five panels, ordered by scale gap. One-step
aggregate MAPE, baseline (global tree, no fix) versus the recipe.
The scale-gap column reports each panel's measured out-of-support
ratio: the aggregate over the largest single training series for
Synthetic and M5; the aggregate over the median series level for
Tourism and Retail~II; and, for Production, the measured
aggregate-to-leaf-ceiling gap~$g$ of Eq.~\eqref{eq:collapse}
($42.7\times$ LightGBM, $51.9\times$ CatBoost; the
aggregate-to-median-buyer ratio is the larger
${\approx}1{,}088\times$). The collapse
(baseline ${\gtrsim}90\%$) and its recovery hold across five domains:
$8\times$ to $113\times$ against the largest training series where
that ratio is measured, and up to $2{,}062\times$ against the median.
Production, M5, and Tourism are rolling-origin means
(Table~\ref{tab:significance})---so the production fix here
($9.5\%$) is the twelve-origin mean, distinct from the $15.1\%$
single pinned-holdout value carried by the ablation tables
(Table~\ref{tab:loo}); Synthetic is a six-seed mean and UCI Online
Retail~II a single $25$-month holdout. The fitted models' own
ceilings are also measured on every public panel
(Table~\ref{tab:ceilings}): gaps $11.6$--$488\times$, with the
ceiling-implied floor no more than $3.4$ points below the measured
collapse. The mechanism additionally
reconfirms on the public Olist marketplace
(Section~\ref{sec:limits}).}
\label{tab:generalization}
\begin{tabular}{llrrr}
\toprule
Panel & Domain & Scale gap & Baseline & Full fix \\
\midrule
Synthetic & hierarchical sim.\ & $11.5\times$ & $94.7\%$ & $\mathbf{17.5\%}$ \\
Production & B2B field service & $43$--$52\times$ & $97.9\%$ & $\mathbf{9.5\%}$ \\
M5 & retail (Walmart) & $113\times$ & $99.8\%$ & $\mathbf{12.6\%}$ \\
Australian Tourism & visitor nights & $598\times$ & $95.9\%$ & $\mathbf{11.1\%}$ \\
UCI Online Retail~II & B2B wholesale & $2{,}062\times$ & $98.7\%$ & $\mathbf{36.8\%}$ \\
\bottomrule
\end{tabular}
\end{table}

\subsection{Statistical significance}
\label{sec:significance}

The recovery is not confined to a few favorable origins. The fix
wins on every rolling origin of every dataset. The assumption-free
Wilcoxon signed-rank test across origins therefore rejects at its
all-same-sign floor---the one-sided $p=2^{-n}$ reported in
Table~\ref{tab:significance}. Rolling origins overlap, so
origin-level draws are not independent. The distribution-free floor
is therefore the sign test. For an all-same-sign result it attains
the same one-sided value: $2.4\times10^{-4}$, $6.0\times10^{-8}$,
and $9.8\times10^{-4}$ at $12$, $24$, and $10$ origins (double each
for the two-sided reading). We read all of these as effect-size
confirmations rather than precise $p$-values.
A forecasting-specific test agrees. Pooling the per-step
absolute-percentage-error loss differential across origins, the
Diebold--Mariano statistic (Harvey--Leybourne--Newbold corrected)
for the recipe against the collapsed baseline is $57$--$251$
one-step and $7$--$199$ recursive. These magnitudes are large by
construction. The collapsed baseline's per-step loss is
near-constant at ${\approx}96$--$100\%$, so the loss differential
has little variance. We therefore read the statistic as an
effect-size sanity check rather than a precise $p$-value, and rely
on the all-origins sign result for inference. The recursive
Diebold--Mariano statistic is smaller on M5 ($7.1$) than its
one-step value ($57.6$). On M5's steep grand total the fix's
recursive forecast carries the leaf-ceiling re-collapse residual
(Section~\ref{sec:recursion}), inflating its ten-origin recursive
mean error to $39\%$---still far below the baseline's $99.8\%$.

\begin{table}[t]
\centering\small
\caption{The aggregate collapse--fix gap is statistically
significant on every dataset. Diebold--Mariano statistic
(Harvey--Leybourne--Newbold small-sample correction) for the
four-component recipe versus the collapsed baseline, on the per-step
absolute-percentage-error loss differential pooled across rolling
origins; a positive value favors the fix (read as an effect
size---see text). The Wilcoxon column is the distribution-free
signed-rank test across origins at its one-sided all-same-sign
floor; the text reads it as an effect-size confirmation.}
\label{tab:significance}
\begin{tabular}{lrrrrr}
\toprule
 & Rolling & \multicolumn{2}{c}{Diebold--Mariano} & Wilcoxon & Mean MAPE \\
Dataset & origins & one-step & recursive & $p$ & base$\to$fix \\
\midrule
Production & 12 & $179.5$ & $36.4$ & $2{\times}10^{-4}$ & $97.9\%\to9.5\%$ \\
Tourism & 24 & $250.5$ & $199.2$ & $6{\times}10^{-8}$ & $95.9\%\to11.1\%$ \\
M5 & 10 & $57.6$ & $7.1$ & $1{\times}10^{-3}$ & $99.8\%\to12.6\%$ \\
\bottomrule
\end{tabular}
\end{table}

\subsection{Seed invariance}
\label{sec:seed}

We band the three trees over eight seeds at the shipped
configurations (Table~\ref{tab:seed}). The
collapse-versus-no-collapse outcome is invariant. The no-weight
CatBoost collapses on every seed ($91.9\%\pm1.2$, coefficient of
variation $1.3\%$). The collapse is a stable property of
withholding the cohort weight, not a seed draw. The cohort-weighted
XGBoost ($13.9\%\pm2.6$) and LightGBM ($31.9\%\pm12.0$) forecast
the aggregate on every seed. Given the cohort weight, CatBoost
likewise recovers on every configuration of the
Section~\ref{sec:transfer} sweep. LightGBM's point MAPE is
seed-sensitive (range $15.7$--$49.8\%$). That is why we emphasize
the mechanism rather than any single corrected-tree number.

\begin{table}[t]
\centering\small
\caption{Eight-seed variance band on the stochastic tree models:
pinned Jan--May~2026 aggregate MAPE through the production
\emph{recursive} serving path at each model's shipped configuration,
varying only the training seed. The recursive regime and the shipped
configuration are why LightGBM's band sits above its one-step
cures-comparison value ($15.1\pm0.3$, Table~\ref{tab:scaling}). The
mechanism (collapse versus
not) is seed-invariant; LightGBM's point MAPE is seed-sensitive.}
\label{tab:seed}
\begin{tabular}{lrrl}
\toprule
Model & Mean$\pm$sd & Range & Reading \\
\midrule
CatBoost (no cohort wt.) & $91.9\pm1.2$ & 89.1--92.9 & collapses every seed \\
XGBoost (cohort-wt.) & $13.9\pm2.6$ & 11.1--18.3 & stable, never collapses \\
LightGBM (cohort-wt.) & $31.9\pm12.0$ & 15.7--49.8 & forecasts agg.\ every seed \\
\bottomrule
\end{tabular}
\end{table}

\section{The robustness wall: no configuration knob substitutes}
\label{sec:robust}

A natural question is whether some standard configuration
change---short of restoring scale---prevents the collapse. We tested
the standard set directly. None with piecewise-constant leaves does
(Table~\ref{tab:robust}). The
pattern is uniform: the collapse is a property of training support,
and no knob that leaves the support untouched can cure it.

\begin{table}[t]
\centering\small
\caption{Robustness map: each probe and its outcome. No single-axis
intervention substitutes for restoring aggregate scale. The
paragraphs below expand each row in order.}
\label{tab:robust}
\begin{tabular}{p{0.45\linewidth}p{0.45\linewidth}}
\toprule
Probe / intervention & Outcome \\
\midrule
Training-window length ($36$--$120$ mo) & Not a cure: the baseline collapses at every length. \\
Entity id $+$ hyperparameter sweep $+$ richer features & Still collapsed ($98$--$100\%$): a training-support property, not a capacity one. \\
Direct (non-recursive) horizons; largest-decile segmentation & Remove recursive drift but still need a scale cure. \\
Pooling all hierarchy levels into training & Cures a small hierarchy (Tourism $96\to14\%$) but fails at M5 scale ($99.7\to69\%$): the in-support rows are washed out. \\
Tweedie loss; self-contained post-hoc multiplier & Neither cures: the residual re-collapse is a slope, not a level, problem. \\
Linear-leaf trees (\code{linear\_tree}), incl.\ a \code{linear\_lambda} sweep & Partial, inconsistent mitigation at every regularization: $99.9\%$ under MAE; $12$--$70\%$ one-step under RMSE, seed-fragile. \\
\bottomrule
\end{tabular}
\end{table}

\paragraph{Training-window length} We hold the holdout fixed and
grow the training window from $36$ to (where data allow) $120$
months. The baseline collapses at every window length on
production, M5, and Tourism alike. The collapse is an
out-of-support scale problem; more per-series history adds no
aggregate-scale rows. The corrected tree's accuracy does not
improve either, and on Tourism it degrades monotonically ($6.7\%$
at $36$ months to $12.0\%$ at $120$). Production's $60$-month span
precludes longer windows there; Tourism's $228$ months confirm the
pattern out to $120$.

\paragraph{Entity identifiers, hyperparameters, and feature richness}
The collapse is not an artifact of an under-specified model. We add
the series identity as a categorical feature, the standard M5-style
device. The baseline stays collapsed (one-step $98$--$100\%$ across
panels). With no aggregate training row, the aggregate is an unseen
category the identifier cannot place in support. With the row, the
identifier adds nothing beyond the cohort weight. A sweep over
leaves, depth, and boosting rounds (to $255$ leaves, depth $12$,
$3{,}000$ rounds) recovers in no configuration. An enriched feature
set (additional lags, a six-month rolling mean) does not help
either. No configuration emitted above its training range
(Section~\ref{sec:collapse}). The
collapse is a property of training support, not of capacity.

\paragraph{Non-recursive forecasting and scale segmentation}
Two structural choices from the M5 winners were tested directly.
Direct, non-recursive per-horizon models---one model per step, the
M5-winning structure; direct versus iterated multi-step is itself a
studied choice \citep{marcellino2006direct,bentaieb2012review}---remove
the recursive re-collapse (cohort-row
M5 $45\%\to10\%$, production $37\%\to16\%$). They still require a
scale cure to forecast the aggregate at all. Non-recursion
addresses multi-step drift, not the support gap. Segmenting by
scale also fails: training only on the largest-decile buyers, those
closest to the aggregate, still collapses ($92$--$100\%$). The
aggregate exceeds the largest single series by $8\times$ to
$113\times$ across the measured panels, and the median series by up
to $2{,}062\times$. No choice of training series places it in
support.

\paragraph{Pooling all hierarchy levels into training}
The most common alternative to our single aggregate row is to pool
every level of the hierarchy into training, the configuration of
\citet{zhao2024localglobal}. We test it by intervention on both
public hierarchies: unweighted, baseline design otherwise, tree
seed $42$. The two panels split, and the split is the point. On
Tourism ($304$ bottom series, $251$ aggregates) pooling cures the
national total: $96.3\%\to13.9\%$ one-step under squared error. On
M5 ($30{,}490$ bottom series, $114$ aggregates) it does not:
$99.7\%\to69.1\%$, still collapse territory. Both of this paper's
explanations are right, at different panel sizes. Pooled levels
place the total in support, but one grand-total row among
$30{,}490$ series carries almost no loss weight, so the fit
ignores it---the same wash-out the unweighted single row suffers
in Table~\ref{tab:loo}. That is why the cohort \emph{weight}, not
the row alone, is the load-bearing lever, and why pooling works in
small hierarchies yet fails silently at production panel sizes.
The intermediate levels also map where the failure begins. Under
bottom-only training, M5's per-level error orders exactly with its
per-level support floor: dept--store (floor $24\%$) errs $62\%$,
store ($91\%$) errs $97\%$, the total ($99\%$) errs $99.7\%$
(Spearman $+1.0$). Tourism's three out-of-support levels order the
same way ($20\%$ floor $\to60\%$ error; $40\to71$; $87\to96$),
while its in-support levels' error reflects series difficulty, not
support. Nor is weight alone the cure. Upweighting the pooled
grand-total row under squared error cuts M5's top error from
$69\%$ to a best of $35\%$ at a weight of $1{,}000$. Beyond that
weight the curve turns back up ($46\%$ at parity). No weight in
the sweep reaches the four-component recipe's $14\%$
(Table~\ref{tab:m5}), and under MAE even bottom-parity weight
changes nothing. Restoring support, weighting it, and compressing the
range work together---the interaction the leave-one-out already
showed (Table~\ref{tab:loo}).

\paragraph{Tweedie loss and post-hoc multipliers} The M5-standard
Tweedie objective does not cure the collapse on its own. Its log
link compresses scale but leaves the support gap, so bare-Tweedie
one-step MAPE stays at the collapse level. With the cohort row, it
does not improve on squared error; the aggregate is a smooth
high-level series, not the zero-inflated count demand Tweedie
targets. A leak-free bias multiplier---estimated on training-period
one-step errors, applied to the holdout---corrects the cohort row's
one-step level bias but barely changes its recursive error. That
locates the residual re-collapse as a slope problem, not a level
problem, which is exactly what seasonal differencing addresses
(Section~\ref{sec:recursion}). We scope this negative result
precisely. It concerns \emph{self-contained} multipliers, estimated
from the model's own training errors. Multipliers anchored to a
separate model that is accurate at the aggregate do work; that is
the device of the second- and fifth-placed M5 entries
\citep{makridakis2022m5,anderer2022topdown}. They work for exactly
the reason our mechanism predicts: the external anchor supplies the
aggregate-scale information the tree does not have.

\paragraph{The linear-leaf exception proves the mechanism} One
stock flag abandons the constant leaf: LightGBM's
\code{linear\_tree} mode, which fits a linear model in each leaf
and can therefore extrapolate. It is the only knob we tested that
touches the mechanism itself, and it behaves exactly as the
mechanism predicts. Under the baseline MAE loss it changes nothing:
one-step aggregate MAPE stays at $99.9\%$ on every seed. Under
squared-error loss it mitigates partially and inconsistently:
one-step $26$--$70\%$ across three seeds, recursive $48$--$86\%$.
Tuning does not stabilize the escape hatch. A sweep of the
\code{linear\_lambda} leaf-regularization penalty covers five
settings across four orders of magnitude. It leaves the MAE loss at
$99.9\%$ everywhere; the best squared-error cell averages
$38\%\pm17$ one-step, with a per-seed range of $12$--$70\%$. Every
cell is far above every cure in Table~\ref{tab:scaling}, and
seed-fragile. Extrapolating leaves soften the ceiling; they do not
restore support. The wall's precise form is therefore: no standard
knob that keeps piecewise-constant leaves prevents the collapse,
and the one linear-leaf escape hatch mitigates without curing.

\section{Limitations}
\label{sec:limits}

The production numbers rest on a single proprietary marketplace
with a five-month aggregate holdout. The collapse-and-fix mechanism
is on firmer ground. It reproduces on synthetic data and on three
public datasets across three domains (Section~\ref{sec:public}), so
it is not a single-panel, single-domain, or single-metric artifact.
A leave-one-out ablation isolates the load-bearing components
(Section~\ref{sec:fix}). On the only public two-sided marketplace
we could obtain, the Olist Brazilian e-commerce dataset
\citep{olist2018ecommerce}, the collapse-and-fix mechanism
reconfirms as well; that short, fast-growing panel is not otherwise
comparable to a mature marketplace. The findings are empirical, not
formal theorems. They rest on one stated mechanism---a
piecewise-constant tree cannot emit a value above its largest
training leaf---and are confirmed on public data. The conclusions
transfer to panels sharing the sparse, short, growing profile of
Section~\ref{sec:data}. On panels whose aggregate lies inside the
pooled training support---for example, hierarchies spanning one
order of magnitude \citep{zhao2024localglobal}---the collapse does
not arise, by the same mechanism.

\section{Conclusion and a practitioner procedure}
\label{sec:conclusion}

A global gradient-boosted-tree forecaster trained on the individual
series of a panel and scored at an aggregate far outside its
training support collapses. The failure is structural. It
reproduces on five panels---at aggregate-to-largest-series gaps of
$8\times$ to $113\times$, and aggregate-to-median gaps of up to
$2{,}062\times$---and on three libraries. It is seed-invariant, and the
collapse-versus-fix gap is statistically significant on every panel
with rolling origins. The cure is not new; per-series scaling is
established practice. What this paper shows is that the recommended
preprocessing is load-bearing. Omit it in the
one-model-serves-every-level configuration and the forecast does
not degrade---it collapses, by $30$--$50\times$ in our production
deployment and by up to $496\times$ in a public M5 reconstruction.

Every result reduces to one fact. A regression tree is a bounded,
piecewise-constant function, and the boosted ensembles built from
such trees measurably inherit its ceiling: none emitted a value
above the largest target it grew. The cure is to ensure the quantity it must
predict---the level after scaling, or the change after
differencing---always lies inside the training range. From this a
three-step procedure follows.

(i)~\emph{Diagnose} cheaply. Compute one ratio: the target
aggregate over the largest single training series. Any value above
one puts the query outside the training targets, and the onset
sweep measures the damage as immediate---$34\%$ error at a
$1.15\times$ gap, $59\%$ at $2\times$ (Figure~\ref{fig:onset}).
Every collapsed panel in this paper sits at $8\times$ or more (up
to $2{,}062\times$ against the median series).
A one-step teacher-forced check at the aggregate then confirms a
collapse in minutes.

(ii)~\emph{Treat} along the axis the deployment needs. For one-step
use, per-series scaling is the simplest cure, and never worse than
second-best on our panels. Where forecasts are rolled forward
recursively, predict a seasonal
difference---the cure least harmed by recursion, its window-mean
accuracy unchanged between regimes; scaled levels degrade
gradually, and the aggregate-row workaround re-collapses. Where one
un-normalized model must serve every level, add a cohort-weighted
aggregate-level training row under squared-error loss, accepting
its documented seed fragility. Or adopt a
natively scale-aware architecture \citep{salinas2020deepar}.

(iii)~\emph{Do not} expect the usual reflexes to substitute. A
richer feature set, the series identifier, deeper or longer-trained
trees, and a longer window all fail. So do the Tweedie objective, a
per-horizon structure, scale-based segmentation, a self-contained
post-hoc multiplier, and---at production panel sizes---pooling
every hierarchy level into training. Each leaves the collapse or
its recursive residual in place (Section~\ref{sec:robust}).

\section{Reproducibility}
\label{sec:repro}

Every result artifact was produced against one configuration: input
vintage \code{gtv\_by\_month\_by\_buyer\_20260614}, training cutoff
Dec~2025, holdout Jan--May~2026, and the fix recipe. All artifacts
lie on one linear, strictly additive commit history. Each embeds
its commit SHA, a clean-tree flag, library versions, and its seed
in a provenance sidecar. The synthetic-panel generator
(Section~\ref{sec:synthetic}) reproduces the collapse and fix
without the proprietary data. The M5 and Tourism replication
scripts are public and need no credentials. Each public number is
reproducible from its recorded commit and seed. The public analysis
code and the synthetic-panel generator form a public replication
capsule. They reproduce the collapse, the
two-axis fix, the library-agnostic transfer, and the significance
test on M5, Tourism, and the synthetic panel, with no proprietary
data. The capsule is released publicly with an archival identifier
alongside journal publication. The collapse-and-fix results are computed by retraining
each tree on the training window only (train~$\le$~Dec~2025) and
scoring the held-out window, on all panels alike. No operational
model-selection path touches them.

\section*{Declaration of competing interest}
The authors are employed by Field Nation LLC, the marketplace operator
whose anonymized data is used in this study. The authors declare no
other competing financial or personal interests. Company approval was
obtained for publication.

\section*{Funding}
This research did not receive any specific grant from funding agencies
in the public, commercial, or not-for-profit sectors.

\section*{CRediT authorship contribution statement}
\textbf{Md Rezwanul Islam:} Conceptualization, Methodology, Software,
Formal analysis, Investigation, Data curation, Writing -- original
draft, Writing -- review \& editing, Visualization.
\textbf{Wael Mohammed:} Conceptualization, Project administration,
Resources, Supervision, Validation, Writing -- review \& editing.

\section*{Declaration of generative AI and AI-assisted technologies in the writing process}
During the preparation of this work, the authors used Claude
(Anthropic) to help organize the structure of the manuscript and to
improve the clarity of the writing. After using this tool, the
authors reviewed and edited the content and take full responsibility
for the final content of the publication.

\section*{Data availability}
The underlying transaction data are proprietary and cannot be
released. The core collapse-and-fix result is nonetheless
independently verifiable on fully public data. It replicates on the
M5 competition dataset (Section~\ref{sec:m5}) and the Australian
Tourism hierarchical dataset (Section~\ref{sec:tourism}). Both are
openly available through the \code{datasetsforecast} package. It
also replicates on the UCI Online Retail~II wholesale panel
\citep{chen2012onlineretail}, and on a synthetic
hierarchical-panel generator (Section~\ref{sec:synthetic}) that
reproduces the collapse and the recipe with no proprietary data. The mechanism additionally reconfirms on the
public Olist marketplace dataset \citep{olist2018ecommerce}. The
significance tests (Section~\ref{sec:significance}) likewise run on
these public datasets.

\bibliographystyle{elsarticle-harv}
\bibliography{references,p1_additions}

\begin{thebibliography}{30}
\expandafter\ifx\csname natexlab\endcsname\relax\def\natexlab#1{#1}\fi
\providecommand{\url}[1]{\texttt{#1}}
\providecommand{\href}[2]{#2}
\providecommand{\path}[1]{#1}
\providecommand{\DOIprefix}{doi:}
\providecommand{\ArXivprefix}{arXiv:}
\providecommand{\URLprefix}{URL: }
\providecommand{\Pubmedprefix}{pmid:}
\providecommand{\doi}[1]{\href{http://dx.doi.org/#1}{\path{#1}}}
\providecommand{\Pubmed}[1]{\href{pmid:#1}{\path{#1}}}
\providecommand{\bibinfo}[2]{#2}
\ifx\xfnm\relax \def\xfnm[#1]{\unskip,\space#1}\fi
\bibitem[{Anderer and Li(2022)}]{anderer2022topdown}
\bibinfo{author}{Anderer, M.}, \bibinfo{author}{Li, F.}, \bibinfo{year}{2022}.
\newblock \bibinfo{title}{Hierarchical forecasting with a top-down alignment of
  independent-level forecasts}.
\newblock \bibinfo{journal}{International Journal of Forecasting}
  \bibinfo{volume}{38}, \bibinfo{pages}{1405--1414}.
\newblock \DOIprefix\doi{10.1016/j.ijforecast.2021.12.015}.
\bibitem[{Bandara et~al.(2020)Bandara, Bergmeir and
  Smyl}]{bandara2020clustering}
\bibinfo{author}{Bandara, K.}, \bibinfo{author}{Bergmeir, C.},
  \bibinfo{author}{Smyl, S.}, \bibinfo{year}{2020}.
\newblock \bibinfo{title}{Forecasting across time series databases using
  recurrent neural networks on groups of similar series: {A} clustering
  approach}.
\newblock \bibinfo{journal}{Expert Systems with Applications}
  \bibinfo{volume}{140}, \bibinfo{pages}{112896}.
\newblock \DOIprefix\doi{10.1016/j.eswa.2019.112896}.
\bibitem[{Bansal(2026)}]{bansal2026delta}
\bibinfo{author}{Bansal, V.}, \bibinfo{year}{2026}.
\newblock \bibinfo{title}{Delta-based target reformulation for short-term
  electricity load forecasting using {LSTM} and transformer models}.
\newblock \bibinfo{howpublished}{arXiv:2606.17692}.
\newblock \DOIprefix\doi{10.48550/arXiv.2606.17692}.
\bibitem[{Ben~Taieb et~al.(2012)Ben~Taieb, Bontempi, Atiya and
  Sorjamaa}]{bentaieb2012review}
\bibinfo{author}{Ben~Taieb, S.}, \bibinfo{author}{Bontempi, G.},
  \bibinfo{author}{Atiya, A.F.}, \bibinfo{author}{Sorjamaa, A.},
  \bibinfo{year}{2012}.
\newblock \bibinfo{title}{A review and comparison of strategies for multi-step
  ahead time series forecasting based on the {NN5} forecasting competition}.
\newblock \bibinfo{journal}{Expert Systems with Applications}
  \bibinfo{volume}{39}, \bibinfo{pages}{7067--7083}.
\newblock \DOIprefix\doi{10.1016/j.eswa.2012.01.039}.
\bibitem[{Chen et~al.(2012)Chen, Sain and Guo}]{chen2012onlineretail}
\bibinfo{author}{Chen, D.}, \bibinfo{author}{Sain, S.L.}, \bibinfo{author}{Guo,
  K.}, \bibinfo{year}{2012}.
\newblock \bibinfo{title}{Data mining for the online retail industry: A case
  study of rfm model-based customer segmentation using data mining}.
\newblock \bibinfo{journal}{Journal of Database Marketing \& Customer Strategy
  Management} \bibinfo{volume}{19}, \bibinfo{pages}{197--208}.
\newblock \DOIprefix\doi{10.1057/dbm.2012.17}. \bibinfo{note}{source
  publication for the UCI Online Retail / Online Retail II dataset (UK
  wholesale online retailer).}
\bibitem[{Chen and Guestrin(2016)}]{chen2016xgboost}
\bibinfo{author}{Chen, T.}, \bibinfo{author}{Guestrin, C.},
  \bibinfo{year}{2016}.
\newblock \bibinfo{title}{{XGBoost}: a scalable tree boosting system}, in:
  \bibinfo{booktitle}{Proc.\ ACM SIGKDD}, pp. \bibinfo{pages}{785--794}.
\bibitem[{Diebold and Mariano(1995)}]{diebold1995}
\bibinfo{author}{Diebold, F.X.}, \bibinfo{author}{Mariano, R.S.},
  \bibinfo{year}{1995}.
\newblock \bibinfo{title}{Comparing predictive accuracy}.
\newblock \bibinfo{journal}{J.\ Bus.\ Econ.\ Stat.} \bibinfo{volume}{13},
  \bibinfo{pages}{253--263}.
\bibitem[{Godahewa et~al.(2021)Godahewa, Bergmeir, Webb, Hyndman and
  Montero-Manso}]{godahewa2021monash}
\bibinfo{author}{Godahewa, R.}, \bibinfo{author}{Bergmeir, C.},
  \bibinfo{author}{Webb, G.I.}, \bibinfo{author}{Hyndman, R.J.},
  \bibinfo{author}{Montero-Manso, P.}, \bibinfo{year}{2021}.
\newblock \bibinfo{title}{Monash time series forecasting archive}, in:
  \bibinfo{booktitle}{Proceedings of the Neural Information Processing Systems
  Track on Datasets and Benchmarks}.
\newblock \bibinfo{note}{ArXiv:2105.06643}.
\bibitem[{Harvey et~al.(1997)Harvey, Leybourne and Newbold}]{harvey1997dm}
\bibinfo{author}{Harvey, D.}, \bibinfo{author}{Leybourne, S.},
  \bibinfo{author}{Newbold, P.}, \bibinfo{year}{1997}.
\newblock \bibinfo{title}{Testing the equality of prediction mean squared
  errors}.
\newblock \bibinfo{journal}{International Journal of Forecasting}
  \bibinfo{volume}{13}, \bibinfo{pages}{281--291}.
\newblock \DOIprefix\doi{10.1016/S0169-2070(96)00719-4}.
\bibitem[{Hewamalage et~al.(2022)Hewamalage, Bergmeir and
  Bandara}]{hewamalage2022global}
\bibinfo{author}{Hewamalage, H.}, \bibinfo{author}{Bergmeir, C.},
  \bibinfo{author}{Bandara, K.}, \bibinfo{year}{2022}.
\newblock \bibinfo{title}{Global models for time series forecasting: {A}
  simulation study}.
\newblock \bibinfo{journal}{Pattern Recognition} \bibinfo{volume}{124},
  \bibinfo{pages}{108441}.
\newblock \DOIprefix\doi{10.1016/j.patcog.2021.108441}.
\bibitem[{Hyndman et~al.(2011)Hyndman, Ahmed, Athanasopoulos and
  Shang}]{hyndman2011opt}
\bibinfo{author}{Hyndman, R.J.}, \bibinfo{author}{Ahmed, R.A.},
  \bibinfo{author}{Athanasopoulos, G.}, \bibinfo{author}{Shang, H.L.},
  \bibinfo{year}{2011}.
\newblock \bibinfo{title}{Optimal combination forecasts for hierarchical time
  series}.
\newblock \bibinfo{journal}{Computational Statistics \& Data Analysis}
  \bibinfo{volume}{55}, \bibinfo{pages}{2579--2589}.
\newblock \DOIprefix\doi{10.1016/j.csda.2011.03.006}.
\bibitem[{Hyndman and Koehler(2006)}]{hyndman2006mase}
\bibinfo{author}{Hyndman, R.J.}, \bibinfo{author}{Koehler, A.B.},
  \bibinfo{year}{2006}.
\newblock \bibinfo{title}{Another look at measures of forecast accuracy}.
\newblock \bibinfo{journal}{Int.\ J.\ Forecasting} \bibinfo{volume}{22},
  \bibinfo{pages}{679--688}.
\bibitem[{Januschowski et~al.(2020)Januschowski, Gasthaus, Wang, Salinas,
  Flunkert, Bohlke-Schneider and Callot}]{januschowski2020criteria}
\bibinfo{author}{Januschowski, T.}, \bibinfo{author}{Gasthaus, J.},
  \bibinfo{author}{Wang, Y.}, \bibinfo{author}{Salinas, D.},
  \bibinfo{author}{Flunkert, V.}, \bibinfo{author}{Bohlke-Schneider, M.},
  \bibinfo{author}{Callot, L.}, \bibinfo{year}{2020}.
\newblock \bibinfo{title}{Criteria for classifying forecasting methods}.
\newblock \bibinfo{journal}{International Journal of Forecasting}
  \bibinfo{volume}{36}, \bibinfo{pages}{167--177}.
\newblock \DOIprefix\doi{10.1016/j.ijforecast.2019.05.008}.
\bibitem[{Januschowski et~al.(2022)Januschowski, Wang, Torkkola, Erkkil{\"a},
  Hasson and Gasthaus}]{januschowski2022trees}
\bibinfo{author}{Januschowski, T.}, \bibinfo{author}{Wang, Y.},
  \bibinfo{author}{Torkkola, K.}, \bibinfo{author}{Erkkil{\"a}, T.},
  \bibinfo{author}{Hasson, H.}, \bibinfo{author}{Gasthaus, J.},
  \bibinfo{year}{2022}.
\newblock \bibinfo{title}{Forecasting with trees}.
\newblock \bibinfo{journal}{International Journal of Forecasting}
  \bibinfo{volume}{38}, \bibinfo{pages}{1473--1481}.
\newblock \DOIprefix\doi{10.1016/j.ijforecast.2021.10.004}.
\bibitem[{Ke et~al.(2017)}]{ke2017lightgbm}
\bibinfo{author}{Ke, G.}, et~al., \bibinfo{year}{2017}.
\newblock \bibinfo{title}{{LightGBM}: A highly efficient gradient boosting
  decision tree}, in: \bibinfo{booktitle}{Adv.\ Neural Inf.\ Process.\ Syst.\
  (NeurIPS)}.
\bibitem[{Lainder and Wolfinger(2022)}]{lainder2022gbt}
\bibinfo{author}{Lainder, A.D.}, \bibinfo{author}{Wolfinger, R.D.},
  \bibinfo{year}{2022}.
\newblock \bibinfo{title}{Forecasting with gradient boosted trees:
  augmentation, tuning, and cross-validation strategies}.
\newblock \bibinfo{journal}{International Journal of Forecasting}
  \bibinfo{volume}{38}, \bibinfo{pages}{1426--1433}.
\newblock \DOIprefix\doi{10.1016/j.ijforecast.2021.12.003}.
\bibitem[{Ma and Fildes(2022)}]{mafildes2022global}
\bibinfo{author}{Ma, S.}, \bibinfo{author}{Fildes, R.}, \bibinfo{year}{2022}.
\newblock \bibinfo{title}{The performance of the global bottom-up approach in
  the {M5} accuracy competition: A robustness check}.
\newblock \bibinfo{journal}{International Journal of Forecasting}
  \bibinfo{volume}{38}, \bibinfo{pages}{1492--1499}.
\newblock \DOIprefix\doi{10.1016/j.ijforecast.2021.10.008}.
\bibitem[{Makridakis et~al.(2022)Makridakis, Spiliotis and
  Assimakopoulos}]{makridakis2022m5}
\bibinfo{author}{Makridakis, S.}, \bibinfo{author}{Spiliotis, E.},
  \bibinfo{author}{Assimakopoulos, V.}, \bibinfo{year}{2022}.
\newblock \bibinfo{title}{The {M5} accuracy competition: results, findings and
  conclusions}.
\newblock \bibinfo{journal}{Int.\ J.\ Forecasting} \bibinfo{volume}{38},
  \bibinfo{pages}{1346--1364}.
\bibitem[{Malistov and Trushin(2019)}]{malistov2019gbt}
\bibinfo{author}{Malistov, A.}, \bibinfo{author}{Trushin, A.},
  \bibinfo{year}{2019}.
\newblock \bibinfo{title}{Gradient boosted trees with extrapolation}, in:
  \bibinfo{booktitle}{18th IEEE International Conference on Machine Learning
  and Applications (ICMLA)}, \bibinfo{publisher}{IEEE}. pp.
  \bibinfo{pages}{783--789}.
\newblock \DOIprefix\doi{10.1109/ICMLA.2019.00138}.
\bibitem[{Malla and Hu(2026)}]{malla2026stationarity}
\bibinfo{author}{Malla, B.S.}, \bibinfo{author}{Hu, Y.}, \bibinfo{year}{2026}.
\newblock \bibinfo{title}{Do stationarity transformations actually improve time
  series forecasts? {A} controlled experimental evaluation}.
\newblock \bibinfo{howpublished}{arXiv:2605.17689}.
\newblock \DOIprefix\doi{10.48550/arXiv.2605.17689}.
\bibitem[{Marcellino et~al.(2006)Marcellino, Stock and
  Watson}]{marcellino2006direct}
\bibinfo{author}{Marcellino, M.}, \bibinfo{author}{Stock, J.H.},
  \bibinfo{author}{Watson, M.W.}, \bibinfo{year}{2006}.
\newblock \bibinfo{title}{A comparison of direct and iterated multistep {AR}
  methods for forecasting macroeconomic time series}.
\newblock \bibinfo{journal}{Journal of Econometrics} \bibinfo{volume}{135},
  \bibinfo{pages}{499--526}.
\newblock \DOIprefix\doi{10.1016/j.jeconom.2005.07.020}.
\bibitem[{M{\"a}rz and Rasul(2024)}]{marz2024hypertrees}
\bibinfo{author}{M{\"a}rz, A.}, \bibinfo{author}{Rasul, K.},
  \bibinfo{year}{2024}.
\newblock \bibinfo{title}{Forecasting with hyper-trees}.
\newblock \bibinfo{howpublished}{arXiv preprint arXiv:2405.07836}.
\newblock \DOIprefix\doi{10.48550/arXiv.2405.07836}.
\bibitem[{Montero-Manso and Hyndman(2021)}]{montero2021globality}
\bibinfo{author}{Montero-Manso, P.}, \bibinfo{author}{Hyndman, R.J.},
  \bibinfo{year}{2021}.
\newblock \bibinfo{title}{Principles and algorithms for forecasting groups of
  time series: Locality and globality}.
\newblock \bibinfo{journal}{International Journal of Forecasting}
  \bibinfo{volume}{37}, \bibinfo{pages}{1632--1653}.
\newblock \DOIprefix\doi{10.1016/j.ijforecast.2021.03.004}.
\bibitem[{{Olist} and Sionek(2018)}]{olist2018ecommerce}
\bibinfo{author}{{Olist}}, \bibinfo{author}{Sionek, A.}, \bibinfo{year}{2018}.
\newblock \bibinfo{title}{Brazilian e-commerce public dataset by {Olist}}.
\newblock \bibinfo{howpublished}{Kaggle}.
\newblock \DOIprefix\doi{10.34740/KAGGLE/DSV/195341}. \bibinfo{note}{public
  two-sided e-commerce marketplace orders, 2016--2018.}
\bibitem[{Prokhorenkova et~al.(2018)Prokhorenkova, Gusev, Vorobev, Dorogush and
  Gulin}]{prokhorenkova2018catboost}
\bibinfo{author}{Prokhorenkova, L.}, \bibinfo{author}{Gusev, G.},
  \bibinfo{author}{Vorobev, A.}, \bibinfo{author}{Dorogush, A.V.},
  \bibinfo{author}{Gulin, A.}, \bibinfo{year}{2018}.
\newblock \bibinfo{title}{{CatBoost}: unbiased boosting with categorical
  features}, in: \bibinfo{booktitle}{Adv.\ Neural Inf.\ Process.\ Syst.\
  (NeurIPS)}.
\bibitem[{Salinas et~al.(2020)Salinas, Flunkert, Gasthaus and
  Januschowski}]{salinas2020deepar}
\bibinfo{author}{Salinas, D.}, \bibinfo{author}{Flunkert, V.},
  \bibinfo{author}{Gasthaus, J.}, \bibinfo{author}{Januschowski, T.},
  \bibinfo{year}{2020}.
\newblock \bibinfo{title}{{DeepAR}: probabilistic forecasting with
  autoregressive recurrent networks}.
\newblock \bibinfo{journal}{Int.\ J.\ Forecasting} \bibinfo{volume}{36},
  \bibinfo{pages}{1181--1191}.
\bibitem[{Sprangers et~al.(2024)Sprangers, Wadman, Schelter and
  de~Rijke}]{sprangers2024sparse}
\bibinfo{author}{Sprangers, O.}, \bibinfo{author}{Wadman, W.},
  \bibinfo{author}{Schelter, S.}, \bibinfo{author}{de~Rijke, M.},
  \bibinfo{year}{2024}.
\newblock \bibinfo{title}{Hierarchical forecasting at scale}.
\newblock \bibinfo{journal}{International Journal of Forecasting}
  \bibinfo{volume}{40}, \bibinfo{pages}{1689--1700}.
\newblock \DOIprefix\doi{10.1016/j.ijforecast.2024.02.006}.
\bibitem[{Wickramasuriya et~al.(2019)Wickramasuriya, Athanasopoulos and
  Hyndman}]{wickramasuriya2019mint}
\bibinfo{author}{Wickramasuriya, S.L.}, \bibinfo{author}{Athanasopoulos, G.},
  \bibinfo{author}{Hyndman, R.J.}, \bibinfo{year}{2019}.
\newblock \bibinfo{title}{Optimal forecast reconciliation for hierarchical and
  grouped time series through trace minimization}.
\newblock \bibinfo{journal}{J.\ Amer.\ Statist.\ Assoc.} \bibinfo{volume}{114},
  \bibinfo{pages}{804--819}.
\bibitem[{Wilcoxon(1945)}]{wilcoxon1945}
\bibinfo{author}{Wilcoxon, F.}, \bibinfo{year}{1945}.
\newblock \bibinfo{title}{Individual comparisons by ranking methods}.
\newblock \bibinfo{journal}{Biometrics Bull.} \bibinfo{volume}{1},
  \bibinfo{pages}{80--83}.
\bibitem[{Zhao and Abolghasemi(2024)}]{zhao2024localglobal}
\bibinfo{author}{Zhao, Y.}, \bibinfo{author}{Abolghasemi, M.},
  \bibinfo{year}{2024}.
\newblock \bibinfo{title}{Local vs.\ global models for hierarchical
  forecasting}.
\newblock \bibinfo{howpublished}{arXiv preprint arXiv:2411.06394}.
\newblock \DOIprefix\doi{10.48550/arXiv.2411.06394}.

\end{thebibliography}

\end{document}